\documentclass[10pt,journal,compsoc]{IEEEtran-customize}

\ifCLASSOPTIONcompsoc
  \usepackage[nocompress]{cite}
\else
  \usepackage{cite}
\fi
\usepackage{graphicx}
\usepackage{color}
\usepackage{xcolor}
\usepackage{booktabs}
\usepackage{multirow}
\usepackage[hyphens]{url}
\usepackage{hyperref}
\usepackage{caption}
\usepackage{amsmath}
\usepackage{subcaption}
\usepackage{amsfonts, amssymb}
\usepackage{colortbl}
\usepackage{enumitem}
\usepackage{amsmath,amsfonts}
\usepackage{algorithmic}
\usepackage{algorithm}
\usepackage{array}
\usepackage{textcomp}
\usepackage{stfloats}
\usepackage{url}
\usepackage{verbatim}
\usepackage{graphicx}
\usepackage{cite}
\usepackage{pifont}
\usepackage{bbm}
\usepackage[utf8]{inputenc}
\usepackage[T1]{fontenc}
\usepackage{hyperref}
\usepackage{tabularx}
\usepackage{amsfonts}
\usepackage{nicefrac}    
\usepackage{microtype}
\usepackage{xcolor}      
\usepackage{graphicx}
\usepackage[edges]{forest}
\usepackage{tikz}
\usetikzlibrary{positioning, fit}
\usepackage{booktabs}
\usepackage{tabularx}
\usepackage{array}
\usepackage{caption}
\usepackage{colortbl}
\usepackage{diagbox}
\usepackage{threeparttable}
\usepackage{booktabs}

\definecolor{purple}{RGB}{128, 0, 128}
\definecolor{LightRed}{rgb}{1,0.92,0.92}
\definecolor{LightOrange}{rgb}{1,0.95,0.88}
\definecolor{LightYellow}{rgb}{1.0,1.0,0.84}
\definecolor{LightGreen}{rgb}{0.9,1.0,0.88}
\definecolor{LightCyan}{rgb}{0.9,1,1}
\definecolor{LightBlue}{rgb}{0.9,0.94,1}
\definecolor{LightIndigo}{rgb}{0.92,0.9,1}
\definecolor{LightMagenta}{rgb}{0.96,0.86,1}
\definecolor{DirtyWhite}{rgb}{0.96,0.96,0.96}

\DeclareSymbolFont{extraup}{U}{zavm}{m}{n}
\DeclareMathSymbol{\varheart}{\mathalpha}{extraup}{86}
\DeclareMathSymbol{\vardiamond}{\mathalpha}{extraup}{87}
\DeclareMathSymbol{\varclubsuit}{\mathalpha}{extraup}{88}

\begin{document}

\title{Diffusion Models for Time Series Forecasting:\\A Survey}
\renewcommand\thefootnote{\fnsymbol{footnote}}
\author{
        Chen Su$^{{\spadesuit}}$, \hspace{0.1cm}
        Zhengzhou Cai$^{{\Diamond}}$\textsuperscript{\dag}, \hspace{0.1cm}
        Yuanhe Tian$^{{\varheart}}$, \hspace{0.1cm}
        Zhuochao Chang$^{{\spadesuit}}$, \hspace{0.1cm}
        Zihong Zheng$^{{\spadesuit}}$, \hspace{0.1cm}
        Yan Song$^{{\spadesuit}}$
        \\
        \vspace{0.2cm}
        $^{\spadesuit}$University of Science and Technology of China \\
        $^{\Diamond}$Beijing University of Posts and Telecommunications \\
        $^{\varheart}$University of Washington \\
        \vspace{0.1cm}
$^{\spadesuit}$\texttt{suchen4565@mail.ustc.edu.cn}  \hspace{0.1cm}
$^{\Diamond}$\texttt{ai@bupt.edu.cn} \\
$^{\varheart}$\texttt{yhtian@uw.edu} \hspace{0.1cm}
$^{\spadesuit}$\texttt{zc230330@mail.ustc.edu.cn} \\
$^{\spadesuit}$\texttt{leozheng06@mail.ustc.edu.cn} \hspace{0.1cm}
$^{\spadesuit}$\texttt{clksong@gmail.com}
}

\IEEEtitleabstractindextext{%

\begin{abstract}
Diffusion models, initially developed for image synthesis, demonstrate remarkable generative capabilities. 
Recently, their application has expanded to time series forecasting (TSF), yielding promising results.
Existing surveys on time series primarily focus on the application of diffusion models to time series tasks or merely provide model-by-model introductions of diffusion-based TSF models, without establishing a systematic taxonomy for existing diffusion-based TSF models.
In this survey, we firstly introduce several standard diffusion models and their prevalent variants, explaining their adaptation to TSF tasks. 
Then, we provide a comprehensive review of diffusion models for TSF, paying special attention to the sources of conditional information and the mechanisms for integrating this conditioning within the models.
In analyzing existing approaches using diffusion models for TSF, we provide a systematic categorization and a comprehensive summary of them in this survey.
Furthermore, we examine several foundational diffusion models applied to TSF, alongside commonly used datasets and evaluation metrics. 
Finally, we discuss the progress and limitations of these approaches, as well as potential future research directions for diffusion-based TSF. 
Overall, this survey offers a comprehensive overview of recent progress and future prospects for diffusion models in TSF, serving as a valuable reference for researchers in the field.
\end{abstract}

\begin{IEEEkeywords}
Time Series Forecasting, Diffusion Models, Conditional Diffusion, Survey
\end{IEEEkeywords}}

\maketitle
\begingroup
  \renewcommand\thefootnote{\fnsymbol{footnote}} % 让脚注处也显示 *
  \footnotetext[2]{Work done at University of Science and Technology of China.}
\endgroup
\IEEEdisplaynontitleabstractindextext
\IEEEpeerreviewmaketitle

\makeatletter
\def\@IEEEcompsocmakefnmark{\hbox{\normalfont\@thefnmark\ }}
\long\def\@makefntext#1{\parindent 1em\indent\hbox{\@IEEEcompsocmakefnmark}#1}
\makeatother

\makeatletter
\def\@IEEEcompsocmakefnmark{\hbox{\normalfont\@thefnmark.\ }}
\long\def\@makefntext#1{\parindent 1em\indent\hbox{\@IEEEcompsocmakefnmark}#1}
\makeatother

\renewcommand{\thefootnote}{\arabic{footnote}}

\section{Introduction}

\IEEEPARstart{T}{ime} {series forecasting (TSF) is a fundamental task that aims to predict future values of a sequence based on historical data.
It plays a vital role in a wide range of real-world applications, such as energy demand forecasting \cite{hong2016probabilistic,chou2018forecasting,wang2018forecasting}, traffic flow prediction \cite{lippi2013short,li2017diffusion,zheng2020traffic}, and healthcare monitoring \cite{gul2009statistical,fassois2009statistical,sun2018predicting}.
Time series data present diverse structures, ranging from univariate sequences that monitor a single variable over time \cite{wu2021autoformer, zhou2022fedformer, nie2022time}, to multivariate sequences capturing interactions among multiple variables \cite{cirstea2022triformer, zheng2022multivariate, liu2023itransformer}, as well as more complex forms such as spatio-temporal data from geo-distributed sensors \cite{yu2017spatio, zhang2020spatio, cao2020spectral} and multimodal sequences that combine temporal signals with text, images, or metadata \cite{liu2024time, wang2025chattime, zhong2025time}.
TSF approaches undergo significant evolution. 
Early approaches primarily rely on conventional statistical models, such as the autoregressive integrated moving average (ARIMA) model \cite{box2015time,shumway2017arima} and exponential smoothing techniques \cite{gardner1985exponential, hyndman2008forecasting, de2011forecasting}. 
Subsequently, the field shifts towards deep learning-based models. These include recurrent neural networks (RNNs) \cite{connor1994recurrent, tokgoz2018rnn, hewamalage2021recurrent, guo2023multivariate}, convolutional neural networks (CNNs) \cite{borovykh2017conditional, livieris2020cnn, mehtab2022analysis, zhang2023ctfnet}, and Transformers \cite{li2019enhancing, wu2020deep,wei2025wavelet,su2025fusing,su2025text}. 
%wu2021autoformer,zhu2023mr
More recently, large-scale Transformer-based large language models (LLMs) \cite{gruver2023large, rasul2023lag, dooley2023forecastpfn} and foundation models \cite{das2024decoder, woo2024unified, liu2024timer} demonstrate powerful zero-shot learning capabilities for TSF and achieve remarkable performance.
}

Among different deep learning approaches, generative ones rise in recent years with various representative models, 
%In recent years, generative approaches
including variational autoencoders (VAEs) \cite{kingma2013auto, rezende2014stochastic, li2022generative}, generative adversarial networks (GANs) \cite{goodfellow2014generative, radford2015unsupervised, koochali2021if}, and especially diffusion models \cite{sohl2015deep, song2019generative, ho2020denoising, song2020score} that attract growing interest for their ability to capture uncertainty and produce high-quality predictions.
Particularly, diffusion models, as latent variable generators, learn complex data distributions through a two-step cycle of progressive noising followed by reverse denoising.
This paradigm advances computer vision, enabling high-fidelity image synthesis, inpainting, and super-resolution without the adversarial instability of GANs \cite{brock2018large, emami2020spa}.
Motivated by such success, diffusive modeling migrates from vision to a wide range of data modalities \cite{li2022diffusion,liu2023audioldm,tian2024diffusion,tian2024aspect}, including the time series data \cite{rasul2021autoregressive, tashiro2021csdi} and provides a robust generative framework for capturing complex temporal patterns.

\IEEEpubidadjcol

Diffusion models for TSF are commonly formulated as conditional generative models, where the objective is to generate future sequences conditioned on past observations and, optionally, additional contextual information \cite{li2022generative, rasul2021autoregressive, alcaraz2022diffusion}.
During training, Gaussian noise is gradually added to the ground-truth future sequence through a forward diffusion process. 
A neural network guided by the historical context is then trained to reverse this noising process. 
At inference time, the model generates forecasts by iteratively denoising the noise, effectively sampling from the learned conditional distribution. 
Compared to conventional models, diffusion-based approaches offer several key advantages: 
(1) they produce probabilistic forecasts, enabling uncertainty quantification by sampling multiple plausible futures; 
(2) they exhibit stable training behavior and avoid adversarial instability; and 
(3) they offer flexible generative capabilities, allowing seamless integration with various conditioning signals such as covariates, control variables, or multimodal inputs \cite{kollovieh2023predict, yuan2024diffusion, liu2024retrieval}.
Critically, diffusion models demonstrate strong empirical performance, achieving competitive or state-of-the-art performance on various TSF benchmark datasets \cite{shen2024multi, liu2024retrieval, li2025diffusion}. 
Consequently, diffusion models are attracting increasing attention within the TSF research community.
Therefore, it is essential to provide a comprehensive review of the diffusion models for TSF, given that most existing literature reviews of the TSF approaches focus on other approaches \cite{ren2023deep, xu2024survey, corradini2024systematic, irani2025positional}.
Furthermore, existing surveys on diffusion models in time series applications \cite{lin2024diffusion, meijer2024rise, yang2024survey} focus more on how diffusion models are applied to various time series tasks, or provide a model-by-model description of existing diffusion-based TSF studies.
Consequently, a dedicated review is needed that offers a systematic and hierarchical taxonomy to organize and synthesize the current landscape of diffusion-based TSF research.

In this survey, we provide a comprehensive survey of diffusion models for TSF. 
We begin by reviewing the foundation diffusion models, including Denoising Diffusion Probabilistic Model (DDPM) \cite{ho2020denoising}, Denoising diffusion implicit model (DDIM) \cite{song2020denoising}, and their adaptation for TSF tasks incorporating Classifier Guidance \cite{dhariwal2021diffusion} and Classifier‑Free Guidance \cite{ho2022classifier} (Section \ref{sec:foundation}).
We then propose a systematic taxonomy to classify existing diffusion-based TSF approaches along two dimensions: conditioning source (e.g., historical time series and multimodal data) (Section \ref{sec:source}), and condition integration process (e.g., feature-centric condition and diffusion-centric condition) (Section \ref{sec:integration}). 
For each category, we summarize its common traits and highlight representative approaches. 
Furthermore, we review widely used datasets (Section \ref{sec:dataset}) and evaluation metrics (Section \ref{sec:eval}). 
Finally, we present an overall analysis of existing approaches, highlight their current progress and limitations (Section \ref{sec:discussion}), and propose potential directions for future research (Section \ref{sec:direction}).
Through this survey, we aim to comprehensively synthesize recent advances, reveal design patterns, and provide deeper insights into the evolving landscape of diffusion-based TSF.

\begin{figure*}[t]
  \centering
  \includegraphics[width=1\linewidth]{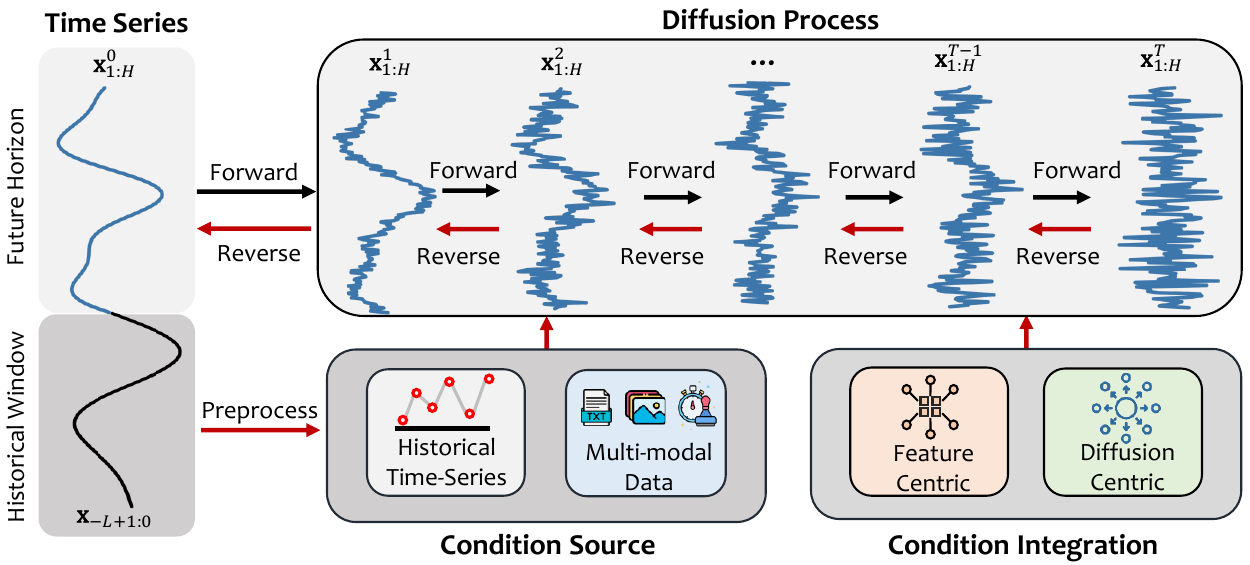}
  \caption{{
General framework of diffusion-based TSF. The left side illustrates the division of time series into a historical window (past time steps) and a future horizon (target sequence). The top-right presents the diffusion process: the forward diffusion process progressively adds Gaussian noise to the clean future time series, while the reverse denoising process reconstructs the future sequence. The bottom-right details the time series conditional diffusion model's core components: Condition Source (Historical Time Series and Multi-modal Data) and Condition Integration (Feature-centric and Diffusion-centric). The black arrows denote the forward diffusion process, and the red arrows represent the model's inference process. Historical time series are preprocessed or combined with multi-modal data to form the condition source, which is then integrated via feature-centric denoising networks or diffusion-centric specialized diffusion processes.
  }}
  \label{fig:Fig.1}
  \vspace{-0.2cm}
\end{figure*}

\section{Foundation Diffusion Models for TSF}
\label{sec:foundation}
{
Standard diffusion models define a generative process that progressively adds noise to data through a forward trajectory and reconstructs the original sample via a reverse denoising process.
Conditional diffusion models\footnote{{Unless otherwise specified, the ``\textit{diffusion model}'' (as well as its specific version, e.g., DDPM \cite{ho2020denoising}) in the following text stand for the conditional diffusion model.}} extend this framework by introducing auxiliary information, such as labels or historical time series, to guide the generation process. 
A representative diffusion model is DDPM \cite{ho2020denoising},
which achieves high-quality generation but requires many iterative sampling steps, resulting in slow inference speed. 
DDIM \cite{song2020denoising} addresses this limitation by reformulating the reverse process as a deterministic trajectory, which accelerates sampling while maintaining compatibility with the original model.
In TSF, the objective is to generate future sequences conditioned on past observations, making the task a natural fit for conditional diffusion models. 
The conditioning input typically consists of historical time series, and the model learns to reconstruct future values from noise under this context. 
Recent research \cite{kollovieh2023predict, feng2024latent, gao2025auto} in diffusion-based TSF also introduces guidance techniques to control the influence of conditioning signals during generation, enabling more flexible and robust forecasting. 
The following subsections provide a detailed roadmap of these concepts, covering the mathematical formulation of DDPM and DDIM, and the adaptation of conditional diffusion models with an emphasis on condition guidance mechanisms for TSF.
}

\subsection{Denoising Diffusion Probabilistic Model (DDPM)}
\label{subsec:ddpm}

{
Denoising Diffusion Probabilistic Models (DDPMs) \cite{ho2020denoising} define a generative process by progressively adding and removing noise through a sequence of forward and reverse steps, conditioned on auxiliary information. 
We denote the original data sample and the conditioning input as $\mathbf{Y}_0$ and $\mathbf{c}$, respectively.
DDPM constructs a forward diffusion process that corrupts $\mathbf{Y}_0$ into a noise distribution and a reverse process that reconstructs $\mathbf{Y}_0$ from noise conditioned on $\mathbf{c}$. 
The following illustrates the details of the forward diffusion process and the reverse denoising process.
}

\textbf{Forward Diffusion Process}
{
The forward process corrupts the data $\mathbf{Y}_0$ using a predefined noise variance schedule $\{\beta_t\}_{t=1}^{T}$, where $\beta_t$ is the noise variance at diffusion step $t$, and $T$ is the total diffusion steps. 
The entire forward process forms a Markov chain that is formulated by
\begin{equation}
q(\mathbf{Y}_{1:T} \mid \mathbf{Y}_0) = \prod_{t=1}^T q(\mathbf{Y}_t \mid \mathbf{Y}_{t-1})
\end{equation}
where $q(\cdot\mid\cdot)$ denotes the forward process distribution and $\mathbf{Y}_t$ is the latent variable at diffusion step $t$. 
In addition, $\mathbf{Y}_{1:T}$ is a noisy data sequence from diffusion step $1$ to $T$. 
Meanwhile, $q(\mathbf{Y}_t \mid \mathbf{Y}_{t-1})=\mathcal{N}(\cdot; \boldsymbol{\mu}, \boldsymbol{\Sigma})$
is a multivariate Gaussian distribution with mean $\boldsymbol{\mu}=\sqrt{1 - \beta_t}\,\mathbf{Y}_{t-1}$ and covariance $\boldsymbol{\Sigma}=\beta_t \mathbf{I}$ (where $\mathbf{I}$ denotes the identity matrix).
The latent variable $\mathbf{Y}_t$ are sampled from a distribution directly conditioned on $\mathbf{Y}_0$:
\begin{equation}
  q(\mathbf{Y}_t \mid \mathbf{Y}_0) = \mathcal{N}(\mathbf{Y}_t; \sqrt{\bar{\alpha}_t} \mathbf{Y}_0, (1 - \bar{\alpha}_t)\mathbf{I})
\end{equation}
where $\alpha_t = 1 - \beta_t$ and $\bar{\alpha}_t = \prod_{i=1}^t \alpha_i$.
The distribution of the endpoint $\mathbf{Y}_T$ of the forward process obeys a Gaussian distribution $\mathcal{N}(\mathbf{0},\mathbf{I})$ with mean $\mathbf{0}$ and covariance $\mathbf{I}$ (which is represented as $\mathbf{Y}_T\sim\mathcal{N}(\mathbf{0},\mathbf{I})$).
}

\textbf{Reverse Denoising Process}
{
Contrary to the forward process, the reverse process reconstructs $\mathbf{Y}_{t-1}$ from $\mathbf{Y}_t$ via a transition distribution $p_\theta(\cdot\mid\cdot)$ parametrized by $\theta$.
This process is formulated as
\begin{equation}
  p_\theta(\mathbf{Y}_{t-1} \mid \mathbf{Y}_t, \mathbf{c}) = \mathcal{N}(\mathbf{Y}_{t-1}; \boldsymbol{\mu}_\theta(\mathbf{Y}_t, \mathbf{c}, t), \Sigma_\theta(\mathbf{Y}_t, \mathbf{c}, t))
\end{equation}
where $\boldsymbol{\mu}_\theta(\cdot)$ is the parameterized mean function and $\Sigma_\theta(\cdot)$ is the parameterized covariance function.
% 在实践中，$\Sigma_\theta(\cdot)$常常被简化为一个仅依赖于$t$的函数。
In practice, $\Sigma_\theta(\cdot)$ is often simplified to $\sigma_t^2\mathbf{I} $ that depends only on $t$.
In training, the model minimizes the KL-divergence between the reverse process $p_\theta(\mathbf{Y}_{t-1} \mid \mathbf{Y}_t, \mathbf{c})$ and the forward process posterior $q(\mathbf{Y}_{t-1} \mid \mathbf{Y}_t, \mathbf{Y}_0)$.
According to Ho et al. \cite{ho2020denoising}, this objective is simplified to a noise prediction loss $\mathcal{L}$ computed by
\begin{equation}
  \mathcal{L} = \mathbb{E}_{\mathbf{Y}_0, \mathbf{c}, t, \boldsymbol{\epsilon}} \left[ \left\| \boldsymbol{\epsilon} - \boldsymbol{\epsilon}_\theta(\mathbf{Y}_t, \mathbf{c}, t) \right\|^2 \right]
\end{equation}
where $\boldsymbol{\epsilon}\sim\mathcal{N}(\mathbf{0},\mathbf{I})$ is the Gaussian noise added to the clean data sample in the forward process, and $\boldsymbol{\epsilon}_\theta(\cdot)$ is the noise prediction network with conditional $\mathbf{c}$ as input. In addition, $\mathbb{E}[\cdot]$ denotes the expectation operation and $\left\| \cdot \right\|^2$ represents Euclidean norm function.
During inference, the model iteratively applies the conditional reverse transitions starting from $\mathbf{Y}_T\sim\mathcal{N}(\mathbf{0},\mathbf{I})$ as
\begin{equation}
\mathbf{Y}_{t-1} \sim p_\theta(\mathbf{Y}_{t-1} \mid \mathbf{Y}_t, \mathbf{c}), \quad t=T,T-1,\dots,1
\end{equation}
Among them, $\widehat{\mathbf{Y}}_0$ sampled at step $t=1$ is the final result of prediction based on the condition $\mathbf{c}$.
}

\subsection{Denoising Diffusion Implicit Model (DDIM)}
\label{subsec:ddim}
{
DDPMs require sampling over hundreds to thousands of total diffusion steps to produce high-quality samples, resulting in prolonged inference times.
Accelerating this sampling process, denoising diffusion implicit models (DDIMs) \cite{song2020denoising} reformulate it as a non-Markovian deterministic process. 
This reformulation permits skipping steps, enabling the attainment of equivalent generation quality using substantially fewer sampling steps than the total diffusion steps. 
According to Song et al. \cite{song2020denoising}, DDIMs share the identical forward process with DDPMs. 
Consequently, DDIMs sampling becomes directly applicable to pre-trained DDPMs without requiring retraining.
This property provides a practical path for deploying diffusion models efficiently.
Specifically, DDIMs reformulates the reverse process as
\begin{equation}
\label{ddim_sampling}
  \mathbf{Y}_{t-1} = \sqrt{\bar{\alpha}_{t-1}} \cdot {\mathbf{Y}}_\theta + \sqrt{1 - \bar{\alpha}_{t-1}} \cdot \boldsymbol{\epsilon}_\theta(\mathbf{Y}_t, \mathbf{c}, t)
\end{equation}
where ${\mathbf{Y}}_\theta$ is computed by
\begin{equation}
  {\mathbf{Y}}_\theta = \frac{1}{\sqrt{\bar{\alpha}_t}} \left( \mathbf{Y}_t - \sqrt{1 - \bar{\alpha}_t} \cdot \boldsymbol{\epsilon}_\theta(\mathbf{Y}_t, \mathbf{c}, t) \right)
\end{equation}
Compared to DDPMs, DDIMs eliminate stochasticity in the reverse process by removing the random noise term at each step, while preserving the marginal distributions of the original diffusion trajectory. 
The step-skip sampling in DDIMs eliminates the need to iterate over all $T$ time steps in the reverse denoising process.
Instead, DDIM samples only on a subset of diffusion steps $t_{s_1}, t_{s_2}, \dots, t_{s_S}$, where $t_{s_1} = T$ and $t_{s_S} = 0$.
Specifically, for sampling step $k$ from $1$ to $S-1$, the DDIMs computes $\mathbf{Y}_{t_{s_{k-1}}}$ from $\mathbf{Y}_{t_{s_k}}$ using Eq. (\ref{ddim_sampling}) and replaces $t$ and $t-1$ with $t_{s_{k}}$ and $t_{s_{k-1}}$, respectively.
}

Beyond DDIMs, several advanced sampling techniques are proposed to accelerate diffusion models. 
For instance, Analytic-DPM \cite{bao2022analytic} leverages closed-form solutions to the reverse diffusion ordinary differential equation, enabling direct computation of sample trajectories without iterative simulation and thus improving the speed.
DPM-Solver \cite{lu2022dpm} applies high-order numerical solvers to the reverse process, replacing the standard Euler approach with more accurate and efficient integration schemes. 
These approaches further reduce inference time and improve sample quality by optimizing the computation of the reverse trajectory.

\begin{figure*}[t]
  \centering
  \includegraphics[width=1\linewidth]{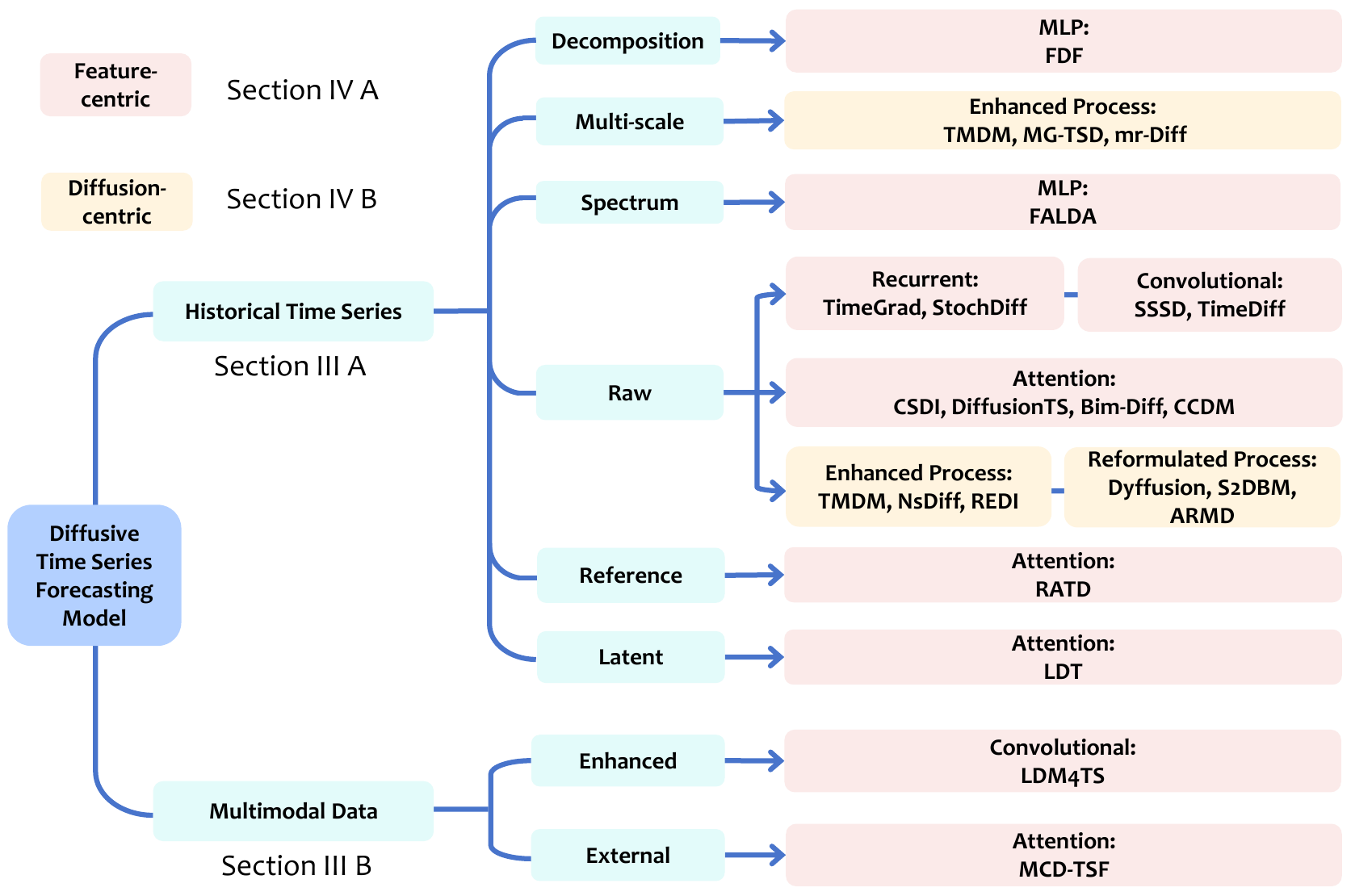}
  \caption{{
  Taxonomy of diffusion-based models for TSF. All approaches are firstly categorized by the condition source, leading to models using historical time series and models using multimodal data.
  The models in each category are further categorized into different sub-categories.
  Meanwhile, we group these models by the condition integration process into feature-centric and diffusion-centric strategies, where feature-centric and diffusion-centric approaches are highlighted in the red and yellow background colors, respectively.
  We also present the section number to illustrate models in different categories.
  }}
  \label{fig:category}
  \vspace{-0.2cm}
\end{figure*}

\subsection{Applying Diffusion Models for TSF}
{
A straightforward approach to applying diffusion models to TSF is to set the original data $\mathbf{Y}_0$ as the future time series $\mathbf{x}_{1:H} = (x_1, x_2, \ldots, x_i, \ldots x_H)$, where $H$ is the prediction horizon and $x_i$ denotes the time series value at timestep $i$.
Meanwhile, the condition $\mathbf{c}$ in diffusion model equals to the historical time series $\mathbf{x}_{-L+1:0} = (x_{-L+1}, x_{-L+2}, \ldots, x_0)$ with $L$ denoting the length of the history time series.
In this formulation, the model learns the conditional distribution $q(\mathbf{x}_{1:H}\mid \mathbf{x}_{-L+1:0})$ of future values given historical time series $\mathbf{x}_{-L+1:0}$. 
The forward diffusion process corrupts the future time series $\mathbf{x}_{1:H}$ with Gaussian noise over $T$ steps, while the reverse process reconstructs $\mathbf{x}_{1:H}$ from noise $\mathbf{x}^T_{1:H}\sim\mathcal{N}(\mathbf{0},\mathbf{I})$, conditioned on the historical time series $\mathbf{x}_{-L+1:0}$.
Specifically, the DDPM for TSF, as illustrated in Fig. \ref{fig:Fig.1}, defines the forward process as
\begin{equation}
q(\mathbf{x}_{1:H}^{1:T} \mid \mathbf{x}_{1:H}^0) = \prod_{t=1}^T q(\mathbf{x}_{1:H}^t \mid \mathbf{x}_{1:H}^{t-1})
\end{equation}
where $\mathbf{x}_{1:H}^t$ is the latent variable at diffusion step $t$ and the forward progress transition is formulated as
\begin{equation}
q(\mathbf{x}_{1:H}^t \mid \mathbf{x}_{1:H}^{t-1}) = \mathcal{N}(\mathbf{x}_{1:H}^t; \sqrt{1 - \beta_t}\,\mathbf{x}_{1:H}^{t-1}, \beta_t \mathbf{I})
\end{equation}
Meanwhile, the reverse process $p_\theta(\mathbf{x}_{1:H}^{t-1} \mid \mathbf{x}_{1:H}^t, \mathbf{x}_{-L+1:0})$ is parameterized as
\begin{equation}
  \mathcal{N} \big( \mathbf{x}_{1:H}^{t-1}; \boldsymbol{\mu}_\theta(\mathbf{x}_{1:H}^t, \mathbf{x}_{-L+1:0}, t), 
  \Sigma_\theta(\mathbf{x}_{1:H}^t, \mathbf{x}_{-L+1:0}, t) \big)
\end{equation}
The training objective is to minimize
\begin{equation}
\mathcal{L} = \mathbb{E}_{\mathbf{x}_{1:H}^0, \mathbf{x}_{-L+1:0}, t, \boldsymbol{\epsilon}} \left[ \left\| \boldsymbol{\epsilon} - \boldsymbol{\epsilon}_\theta(\mathbf{x}_{1:H}^t, \mathbf{x}_{-L+1:0}, t) \right\|^2 \right]
\end{equation}
The DDIM variant for TSF also modifies the reverse process to a deterministic trajectory, which is formulated as
\begin{equation}
\mathbf{x}_{1:H}^{t-1} = \sqrt{\bar{\alpha}_{t-1}} \cdot {\mathbf{x}}_\theta + \sqrt{1 - \bar{\alpha}_{t-1}} \cdot \boldsymbol{\epsilon}_\theta(\mathbf{x}_{1:H}^t, \mathbf{x}_{-L+1:0}, t)
\end{equation}
where ${\mathbf{x}}_\theta$ is derived from the predicted noise and the conditioning input:
\begin{equation}
  {\mathbf{x}}_\theta = \frac{1}{\sqrt{\bar{\alpha}_t}} \left( \mathbf{x}^t_{1:H} - \sqrt{1 - \bar{\alpha}_t} \cdot \boldsymbol{\epsilon}_\theta(\mathbf{x}^t_{1:H}, \mathbf{x}_{-L+1:0}, t) \right)
\end{equation}
}

{
In the field of conditional generation for vision, guidance techniques are proposed to flexibly control the strength of conditioning and improve sample quality.
Recent studies in diffusion-based TSF also adapt guidance mechanisms to incorporate additional conditioning sources without modifying model architectures. 
Guidance mechanisms provide a way to steer the generation process toward desired properties or constraints.
One of these guidance techniques, named classifier guidance, modifies the reverse process by adding the gradient of a classifier\footnote{{Although TSF denotes this component as a predictor, we use the term ``classifier'' to align with the original nomenclature.}} $p_\text{clf}(\mathbf{x}_{-L+1:0})$ trained to predict the historical time series $\mathbf{x}_{-L+1:0}$ from the noisy future sequence $\mathbf{x}_{1:H}^t$. 
According to Dhariwal et al. \cite{dhariwal2021diffusion}, the guidance is simplified to an update to the mean function:
\begin{equation}
\tilde{\boldsymbol{\mu}}_\theta=\boldsymbol{\mu}_\theta(\mathbf{x}_{1:H}^t, \mathbf{x}_{-L+1:0}, t) + s \cdot \nabla_{\mathbf{x}_{1:H}^t} \log p_{\text{clf}}(\mathbf{x}_{-L+1:0} \mid \mathbf{x}_{1:H}^t)
\end{equation}
where $\tilde{\boldsymbol{\mu}}_\theta$ is the updated mean function, $\nabla$ is the gradient operator and $s$ represents guidance scale.
}

{
Different from the classifier guidance that requires additional training of a classifier, classifier-free guidance \cite{ho2022classifier} trains the diffusion model to handle both conditional and unconditional generations by randomly dropping the conditioning input during training. 
In inference, the model combines the conditional and unconditional predictions to update the noise $\boldsymbol{\epsilon}_{\text{guided}}$ by:
\begin{equation}
\boldsymbol{\epsilon}_{\text{guided}} = (1 + s) \cdot \boldsymbol{\epsilon}_\theta(\mathbf{x}_{1:H}^t, \mathbf{x}_{-L+1:0}, t) - s \cdot \boldsymbol{\epsilon}_\theta(\mathbf{x}_{1:H}^t, \varnothing, t)
\end{equation}
where $\boldsymbol{\epsilon}_\theta(\mathbf{x}_{1:H}^t, \varnothing, t)$ is the unconditional noise prediction, $\varnothing$ represents the historical time series is set to a null tensor, and $s \ge 0$ is the guidance scale controlling the strength of conditioning during sampling ($s=0$ yields the conditional prediction, larger $s$ increases guidance).
These mechanisms enable flexible integration of historical and auxiliary information in diffusion-based TSF models.
}

\section{Conditions for TSF Diffusion Models}
\label{sec:source}
{
Based on the conditioning sources for diffusion models in TSF, existing research is categorized into two groups, namely historical time series conditions and multimodal conditions, as illustrated in Fig. \ref{fig:category}.
Approaches that utilize historical time series conditions generally apply processing techniques from the time series domain to decompose complex temporal dependencies into explicit and simpler features, thereby embedding time series priors into the diffusion model.
This feature decomposition reduces the difficulty for the diffusion model to capture temporal correlations and, in certain cases, facilitates more precise uncertainty estimation.
The other category applies multimodal conditioning sources, where the core idea is to introduce information from additional modalities to complementarily describe the time series from different perspectives and thus provide richer contextual information.
In the following subsections, we elaborate on these two main categories of conditioning sources and their specific implementations in diffusion models.
}

\begin{figure}[t]
  \centering
  \includegraphics[width=1\linewidth]{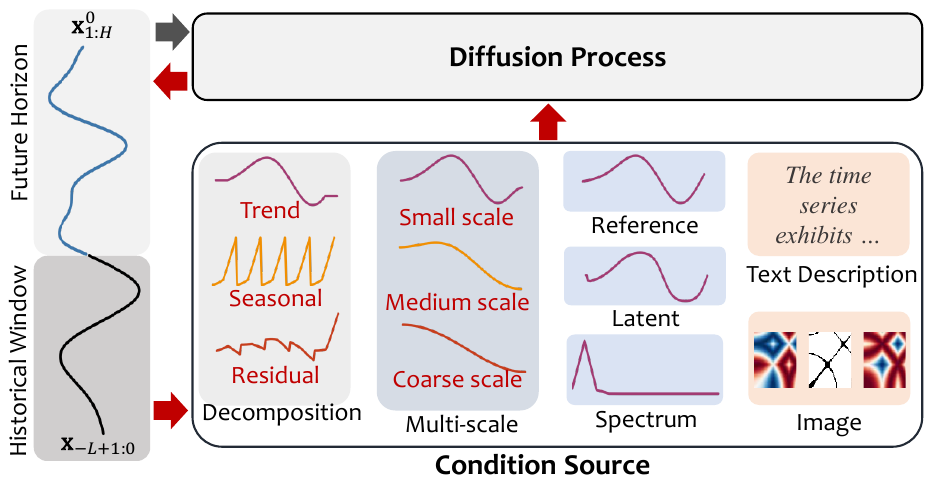}
  \caption{{
  Conditional sources of the diffusion TSF model. The left panel shows historical and prediction windows, the top-right part illustrates a general diffusion process, and the bottom-right presents condition sources, including historical preprocessing and multimodal inputs.
  The grey arrows indicate the forward diffusion process, and the red arrows denote the reverse process.}
  }
  \label{fig:condition_source}
  \vspace{-0.2cm}
\end{figure}

\subsection{Conditioning on Historical Time Series} 
{
The historical time series itself directly serves as the conditioning input for the diffusion model without requiring any additional preprocessing, which we denote as \textit{Raw} \cite{rasul2021autoregressive, tashiro2021csdi, alcaraz2022diffusion, shen2023non, wang2024treating, liu2024stochastic, li2024channel, li2025diffusion}. 
Beyond directly using the raw series, specific time series processing techniques and transformations extract additional information, which injects domain priors as explicit features, as depicted in Fig. \ref{fig:condition_source}. 
According to the employed processing strategy, diffusion-based models conditioning on historical time series are divided into five groups, namely \textit{Multi-scale}, \textit{Decomposition}, \textit{Spectrum}, \textit{Retrieval reference}, and \textit{Latent}.

\textbf{Multi-scale} This strategy prepares multiple resolutions of the same time series so that the condition summarizes dynamics from coarse to fine \cite{fan2024mg, shen2024multi}. 
A generic construction is written as
\begin{equation}
    \mathbf{x}^{(s)}_{-L+1:0}=\big(\mathbf{x}_{-L+1:0}*h^{(s)}\big)_{t s},\qquad s\in\{1,\ldots,S\}
\end{equation}
where $h^{(s)}$ denotes a smoothing kernel at scale $s$, the operator $*$ denotes discrete convolution, the subscript $t s$ denotes subsampling by factor $s$, and $S$ denotes the number of scales. 
The condition aggregates these resolutions as $\mathbf{c}_{\text{ms}}=\{\mathbf{x}^{(1)}_{-L+1:0},\ldots,\mathbf{x}^{(S)}_{-L+1:0}\}$.

\textbf{Decomposition} This strategy separates the series into interpretable components such as trend, seasonality, and residuals \cite{zhang2024fdf, cleveland1990stl}. 
An additive form is formulated as
\begin{equation}
    x_t=T_t+S_t+R_t
\end{equation}
where $x_t$ denotes the observation at time $t$, $T_t$ denotes the trend, $S_t$ denotes the seasonal component with period $P$, and $R_t$ denotes the remainder. The condition stacks $\mathbf{c}_{\text{dec}} = \{T_{-L+1:0}, S_{-L+1:0}, R_{-L+1:0}\}$.
\textbf{Spectrum} This strategy maps the time series to frequency domain features and combines them with time domain features \cite{yuan2024diffusion, wang2025effective}. 
A discrete Fourier transform is written as
\begin{equation}
    \hat{x}_k=\sum_{t=0}^{L-1}x_t\,e^{-i2\pi k t/L},\qquad k=0,\ldots,L-1
\end{equation}
where $\hat{x}_k$ denotes the complex coefficient at frequency index $k$ and $i$ denotes the imaginary unit. The magnitude and phase are $A_k=|\hat{x}_k|$ and $\Phi_k=\arg(\hat{x}_k)$. 
The condition assembles $\mathbf{c}_{\text{spec}}=\{\mathbf{x}_{-L+1:0},A_{0:L-1},\Phi_{0:L-1}\}$ or uses magnitudes only when a phase-free representation is preferred.

\textbf{Retrieval reference} This strategy retrieves similar sequences from a reference set and concatenates them with the target history as auxiliary context \cite{liu2024retrieval}. 
A generic pipeline defines an embedding map $E(\cdot)$ and a cosine similarity $s(\mathbf{u},\mathbf{v})=\frac{\langle E(\mathbf{u}),E(\mathbf{v})\rangle}{|E(\mathbf{u})|_2,|E(\mathbf{v})|_2}$, then selects the top $K$ neighbors $\{\mathbf{u}_{(k)}\}_{k=1}^{K}$ for the query window $\mathbf{x}_{-L+1:0}$. 
A soft weighting $w_k$ is computed by
\begin{equation}
    w_k=\frac{\exp(\tau s_k)}{\sum_{j=1}^{K}\exp(\tau s_j)},\qquad s_k=s(\mathbf{u}_{(k)},\mathbf{x}_{-L+1:0})
\end{equation}
with the temperature $\tau>0$. 
Thus, the reference condition $\mathbf{c}_{\text{ret}}$ is computed through a weighted sum operation by
\begin{equation}
    \mathbf{c}_{\text{ret}}=\sum_{k=1}^{K}w_k\,\mathbf{u}_{(k)}
\end{equation}
and $\mathbf{c}_{\text{ret}}$ has the same temporal length as each neighbor after alignment or interpolation.
\textbf{Latent} This strategy encodes the time series into a compact latent representation that serves as the condition while reducing dimensionality \cite{feng2024latent}. 
A variational encoding is formulated by
\begin{equation}
    q_\phi(\mathbf{z}\mid \mathbf{x}_{-L+1:0})=\mathcal{N}\big(\boldsymbol{\mu}_\phi(\mathbf{x}_{-L+1:0}),\operatorname{diag}(\boldsymbol{\sigma}^2_\phi(\mathbf{x}_{-L+1:0}))\big)
\end{equation}
with sample $\mathbf{z}\sim q_\phi(\mathbf{z}\mid \mathbf{x}_{-L+1:0})$ and condition $\mathbf{c}_{\text{lat}}=\mathbf{z}$,
where $\boldsymbol{\mu}_\phi(\cdot)$ and $\boldsymbol{\sigma}_\phi(\cdot)$ denote encoder outputs parameterized by $\phi$ and $\operatorname{diag}(\cdot)$ denotes a diagonal covariance. 
}

\subsection{Conditioning on Multimodal Data}
{
Multimodal approaches \cite{liu2024time, wang2025chattime, jiang2025multi, zhong2025time} are attracting increasing attention in the TSF domain, and diffusion-based TSF approaches also integrate additional modal information as complementary conditions.
Beyond preprocessing and transformation techniques that extract useful information from historical time series as domain-specific priors, multimodal diffusion-based TSF is to incorporate multimodal data as supplementary conditioning sources for the diffusion model, as depicted in Fig. \ref{fig:condition_source}.
We divide these approaches into \textit{Enhanced} and \textit{External}, based on whether the multimodal data originates from the time series itself or from external sources. 
A unified notation for the multimodal condition {$\mathbf{c}_{\mathrm{mm}}$} uses a single formula:
\begin{equation}
\mathbf{c}_{\mathrm{mm}}=\{\mathbf{x}_{-L+1:0},\mathbf{u}_{\mathrm{mm}}\}
\end{equation}
where $\mathbf{u}_{\mathrm{mm}}$ denotes a multimodal embedding produced by a preprocessing map that aggregates auxiliary signals. 
For \textit{Enhanced} settings, $\mathbf{u}_{\mathrm{mm}}$ derives from the time series, for example, visual latents from rendered images and a text embedding from a template description. 
For \textit{External} settings, $\mathbf{u}_{\mathrm{mm}}$ aggregates aligned side information such as timestamp features and an embedding of external texts.
}

\textbf{Enhanced}
{
The enhanced multimodal model, for example, LDM4TS \cite{ruan2025vision}, derives visual and textual conditions from the same historical time series without using external data. 
The preprocessing pipeline first converts the series into multi-view images that summarize temporal structure from complementary perspectives, for example, a segmentation-based rearrangement that highlights seasonal stripes, a Gramian angular field that encodes pairwise temporal relations as image intensities, and a recurrence plot that marks repeated state patterns. 
These images are then passed through a pretrained vision encoder to obtain compact visual latents aligned with the time series. In parallel, the pipeline constructs a template-based textual description that records dataset identity, forecasting task specification, historical and horizon lengths, basic statistics such as min, max, and median, a coarse trend label, and top autocorrelation lags. Then, a frozen language model encodes this description into a text embedding. 
Therefore, the final conditioning set contains visual latents and a text vector, both deterministically computed from the input time series.
}

\textbf{External}
The external multimodal approach, for example,  
MCD-TSF \cite{su2025multimodal} utilizes side information aligned with the time axis, including timestamp metadata and news text with the target series. 
Preprocessing firstly normalizes calendar attributes, including day of week, month, and holiday flags, into fixed-length timestamp features and aligns them to each historical step. 
It then collects documents that fall within the same interval as the history window, performs standard text cleaning and deduplication, and encodes each document with a pretrained language model. 
The resulting vectors are aggregated to form a single text representation. The multimodal condition is obtained by aligning the numeric history with the timestamp features and the aggregated text vector. 
When text is missing, the conditioning reduces to numeric history plus timestamps, which keeps the pipeline applicable in partially observed settings.

\section{Condition Integration}
\label{sec:integration}
\begin{figure}
    \centering
    \includegraphics[width=\linewidth]{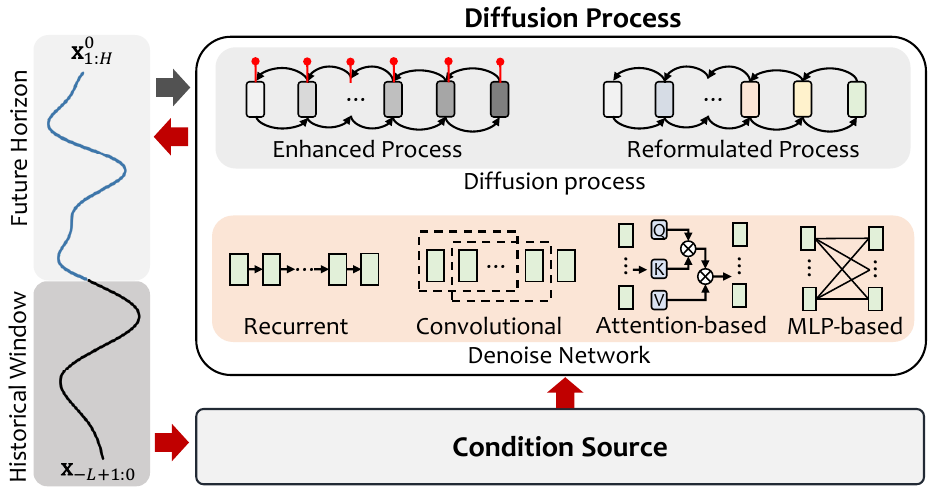}
    \caption{
    {
    Condition integration approaches for diffusion-based TSF models. The left part shows the historical and prediction windows. The bottom right part denotes a generic condition source. The top right part illustrates that the condition firstly enters a denoise network (e.g., a recurrent, convolutional, attention-based, or MLP-based structure) to extract informative features. These features are then incorporated into the diffusion process, which is organized into two categories referred to as enhanced process and reformulated process, representing alternative ways of embedding conditioning information into the generative trajectory.
    }}
    \label{fig:condition_integration}
    \vspace{-0.2cm}
\end{figure}

{
In addition to the source used for conditioning, diffusion-based TSF approaches also differ in their approaches to integrating conditioning information into the generative diffusion process. 
Existing studies integrate time series conditions by designing a specialized denoising network for the diffusion model or modifying the diffusion process.
Therefore, we classify existing condition integration mechanisms into two primary categories, \textit{feature-centric} and \textit{diffusion-centric} approaches, as depicted in Fig. \ref{fig:condition_integration}.
Feature-centric strategies \cite{rasul2021autoregressive, liu2024stochastic, tashiro2021csdi} generally employ the standard diffusion model but rely on a well-designed denoise network to extract diffusion-step-aware features from the historical conditioning data. 
These features then act as input to the learned denoising function, guiding the reverse process. 
Diffusion-centric approaches \cite{li2024transformer, ye2025non, ruhling2023dyffusion} propose modifications to the diffusion process. 
Such modifications incorporate historical data priors directly into both forward and reverse diffusion trajectories, enabling finer‑grained control over the generation.
The following subsections explain the various studies, detailing their methodologies and characteristics.
}

\subsection{Feature-centric Condition}
{
Feature-centric diffusion approaches retain the standard forward noising and reverse denoising processes while placing the design focus on the denoising network as a feature extractor that maps the historical window into conditioning features for every diffusion step. A generic reverse transition for the future horizon $\mathbf{x}_{1:H}$ at step $t$ writes
\begin{equation}
    \mathbf{x}_{1:H}^{t-1}\sim\mathcal{N}\big(\boldsymbol{\mu}_\theta(\mathbf{x}_{1:H}^{t},t,\phi(\mathbf{x}_{-L+1:0})),\boldsymbol{\Sigma}_\theta(\mathbf{x}_{1:H}^{t},t,\phi(\mathbf{x}_{-L+1:0}))\big)
\end{equation}
The function $\phi(\cdot)$ denotes the feature extractor. 
Based on the core model structure of the feature extractor, we categorize feature-centric approaches as \textit{MLP-based}, \textit{Recurrent}, \textit{Convolutional}, and \textit{Attention-based}.
}

\textbf{MLP-based feature extractors}
{
MLP-based approaches \cite{zhang2024fdf, li2025diffusion,wang2025effective} use feed-forward blocks to distill patterns from the history and to modulate conditioning through adaptive layer normalization, which is formulated as
\begin{equation}
\operatorname{AdaLN}(\mathbf{h}, \mathbf{c})=\gamma(\mathbf{c})\odot\frac{\mathbf{h}-\mu(\mathbf{h})}{\sigma(\mathbf{h})}+\beta(\mathbf{c})
\end{equation}
where $\operatorname{AdaLN}(\mathbf{h},\mathbf{c})$ denotes adaptive layer normalization of $\mathbf{h}$ conditioned on $\mathbf{c}$, $\mathbf{h}$ denotes an intermediate feature, $\mathbf{c}$ denotes conditioning features from the history, $\mu(\mathbf{h})$ and $\sigma(\mathbf{h})$ denote feature wise mean and standard deviation, $\odot$ denotes Hadamard product, and $\gamma(\cdot)$ and $\beta(\cdot)$ are MLP outputs driven by $\mathbf{c}$.
FALDA \cite{wang2025effective} performs a light decomposition that splits the signal as $x_t=T_t+S_t+N_t$ with a moving average trend $T_t$, a frequency-selected stationary part $S_t$, and the residual noise $N_t=x_t-T_t-S_t$. 
The decomposed features drive an MLP path under adaptive layer normalization with scale and shift produced from the condition. 
D3U \cite{li2025diffusion} prepares a deterministic forecast $\hat{\mathbf{y}}_{1:H}$ and models residual uncertainty $\mathbf{r}_{1:H}^{0}=\mathbf{y}_{1:H}-\hat{\mathbf{y}}_{1\:H}$ with an MLP based conditional predictor under AdaLN. 
}

\textbf{Recurrent feature extractors}
{
Recurrent approaches \cite{rasul2021autoregressive,liu2024stochastic} leverage hidden state dynamics to propagate historical context through time and across diffusion steps. 
TimeGrad \cite{rasul2021autoregressive} processes the observed sequence with a recurrent cell and maintains a hidden state that updates as
\begin{equation}
    \mathbf{h}_{i}=\operatorname{RNN}(\operatorname{concat}(\mathbf{x}_{i},\mathbf{c}_{i}),\mathbf{h}_{i-1})
\end{equation}
where $\mathbf{x}_i$ denotes the value at time index $i$, $\mathbf{c}_i$ denotes covariates at index $i$, and {
$\operatorname{concat}(\cdot,\cdot)$ concatenates its arguments along the feature (channel) dimension to form the RNN input at step $i$.
}
The denoiser at horizon index $i$ then conditions on $\mathbf{h}_{i-1}$. StochDiff \cite{liu2024stochastic} integrates an LSTM into the feature extractor and defines a data-driven prior at each horizon index, which is formulated as
\begin{equation}
    \mathbf{z}_i\sim\mathcal{N}\big(\boldsymbol{\mu}_\theta(\mathbf{h}_{i-1}),\boldsymbol{\Sigma}_\theta(\mathbf{h}_{i-1})\big)
\end{equation}
\begin{equation}
    \mathbf{h}_{i}=\operatorname{LSTM}(\operatorname{concat}(\mathbf{x}_i,\mathbf{c}_i),\mathbf{h}_{i-1})
\end{equation}
{where} $\mathbf{z}_i$ serves as a prior feature for the reverse step of index $i$ and depends on the previous hidden state $\mathbf{h}_{i-1}$. This design supplies inductive bias for sequential order and regime shifts while remaining compatible with non-autoregressive denoising over the full horizon.
}

\textbf{Convolutional feature extractors}
{
Convolutional approaches \cite{alcaraz2022diffusion,shen2023non,ruan2025vision} rely on one-dimensional convolutions and convolutional sequence kernels to capture temporal dependencies with high efficiency. SSSD \cite{alcaraz2022diffusion} adopts an S4 based convolutional backbone where a learned kernel $K$ induces a causal convolution
\begin{equation}
    \mathbf{y}=(K*\mathbf{z}^{t})=\sum_{\tau\ge 0}K[\tau]\mathbf{z}^{t-\tau}
\end{equation}
where $\mathbf{z}^{t}$ denotes the diffused input at step $t$, $K[\tau]$ denotes the discrete kernel produced by the state space parameterization, and the sum runs over nonnegative lags. 
TimeDiff \cite{shen2023non} employs an encoder and a decoder made of stacked one-dimensional convolutions together with training time conditioning that aligns the boundary between past and future. The future mixup condition uses a stochastic mask $\mathbf{m}_t$ and a projection $\mathcal{F}$ to form
\begin{equation}
    \mathbf{z}_{\text{mix}}=\mathbf{m}_t\odot\mathcal{F}(\mathbf{x}_{-L+1:0})+(\mathbf{1}-\mathbf{m}_t)\odot\mathbf{x}_{1:H}
\end{equation}
{where} $\mathbf{m}_t$ samples a binary mask at diffusion step $t$ and $\mathcal{F}$ denotes a learnable projection of the history into the horizon shape. 
The linear initialization {$\mathbf{z}_{\text{ar}}$ is computed} by
\begin{equation}
    \mathbf{z}_{\text{ar}}=\sum_{i=-L+1}^{0}\mathbf{W}_i\odot\mathbf{X}^0_i+\mathbf{B}
\end{equation}
Here $\mathbf{X}^0_i$ replicates $\mathbf{x}_i$ across $H$ future positions and $\mathbf{W}_i$ and $\mathbf{B}$ denote learnable weights and bias. 
The final condition uses $\mathbf{c}=\operatorname{concat}(\mathbf{z}_{\text{mix}},\mathbf{z}_{\text{ar}})$. LDM4TS \cite{ruan2025vision} converts the series into multi-view images and applies latent diffusion with a convolutional UNet denoiser on image latents. A compact notation writes $ \mathbf{z}_{\text{img}} = E(\mathcal{I}(\mathbf{x}_{-L+1:0})) $ where $\mathcal{I}(\cdot)$ denotes the image transformation and $E(\cdot)$ denotes the vision encoder.
}

\textbf{Attention-based feature extractors}
{
Attention-based approaches \cite{tashiro2021csdi,yuan2024diffusion,feng2024latent,wang2024treating,liu2024retrieval,li2024channel,su2025multimodal} use self-attention to model long-range dependencies over time and across variables and to integrate side information or memories within the denoiser. 
CSDI \cite{tashiro2021csdi} stacks temporal and feature-wise attention to produce conditioning features for imputation and forecasting with
\begin{equation}
    \operatorname{Attn}(\mathbf{Q},\mathbf{K},\mathbf{V})=\operatorname{softmax}\!\big(\mathbf{Q}\mathbf{K}^{\top}/\sqrt{d_k}\big)\mathbf{V}
\end{equation}
\begin{equation}
    \mathbf{Q}=\mathbf{X}\mathbf{W}_Q,\ \mathbf{K}=\mathbf{X}\mathbf{W}_K,\ \mathbf{V}=\mathbf{X}\mathbf{W}_V
\end{equation}
where $\mathbf{Q},\mathbf{K},\mathbf{V}$ are the query, key, and value matrices obtained from the token matrix $\mathbf{X}$ via learned projections $\mathbf{W}_Q,\mathbf{W}_K,\mathbf{W}_V$. 
$\operatorname{Attn}(\cdot)$ denotes scaled dot-product attention, and $\operatorname{softmax}(\cdot)$ performs row-wise normalization to convert similarities into weights over keys for each query. 
$\mathbf{X}$ stacks temporal and variable tokens, and $d_k$ is the key dimension. 
Diffusion-TS \cite{yuan2024diffusion} adopts a Transformer encoder and decoder with decomposition guided representations and augments the diffusion objective with a frequency loss $\mathcal{L}_{\text{freq}}$:
\begin{equation}
    \mathcal{L}_{\text{freq}}=\big\|\mathcal{F}(\hat{\mathbf{x}}_{1:H})-\mathcal{F}(\mathbf{x}_{1:H})\big\|^2
\end{equation}
where $\mathcal{F}(\cdot)$ denotes the discrete Fourier transform and $\hat{\mathbf{x}}_{1:H}$ denotes the model output. 
LDT \cite{feng2024latent} performs diffusion in a compact latent space with a Transformer denoiser and self-conditioning, where the latent at the step $t$ augments the input for the step $t-1$ through
\begin{equation}
    \tilde{\mathbf{z}}^{t-1}=\operatorname{concat}\big(\mathbf{z}^{t-1},g(\mathbf{z}^{t})\big)
\end{equation}
where $\tilde{\mathbf{z}}^{t-1}$ denotes the self-conditioned latent input to the denoiser at step $t-1$ formed by concatenating $\mathbf{z}^{t-1}$ with the projected latent $g(\mathbf{z}^{t})$, $\mathbf{z}^{t}$ denotes the latent variables at step $t$, and $g(\cdot)$ denotes a learned projection for self-conditioning. Bim-Diff \cite{wang2024treating} introduces semantic and episodic memories $\mathbf{M}_s$ and $\mathbf{M}_e$ that produce memory priors through attention
\begin{equation}
    \mathbf{m}_s^j=\sum_i \omega_{s,i}^j\mathbf{M}_{s,i},\qquad \mathbf{m}_e^j=\sum_i \omega_{e,i}^j\mathbf{M}_{e,i}
\end{equation}
where the index $j$ denotes the channel index, $\omega_{s,i}^j$ and $\omega_{e,i}^j$ denote softmax normalized scores computed from queries $\mathbf{h}_j$, and $\mathbf{M}_{s,i}$ and $\mathbf{M}_{e,i}$ denote memory items. The combined prior $\mathbf{m}=\mathbf{m}_s+\mathbf{m}_e$ passes through a projection and a mask to yield mixed condition $\mathbf{c}_{\text{mix}}$:
\begin{equation}
\mathbf{c}_{\text{mix}}=\mathbf{m}_t\odot\mathbf{W}\mathbf{m}+(\mathbf{1}-\mathbf{m}_t)\odot\mathbf{x}_{1:H}
\end{equation}
where $\mathbf{m}_t$ denotes a stochastic mask and $\mathbf{W}$ denotes a learnable projection that maps the memory prior into the condition space. RATD \cite{liu2024retrieval} retrieves neighbor time series ${\mathbf{u}_k}$ from a reference set and injects them by cross attention $\operatorname{Attn(\mathbf{Q},\mathbf{K},\mathbf{V})}$ with $\mathbf{Q}$ from the current hidden representation and with $\mathbf{K}$ and $\mathbf{V}$ from the retrieved set. 
CCDM \cite{li2024channel} combines per-channel MLP encoders with a channel-mixing diffusion Transformer, where a channel-independent dense encoder produces univariate features and a mixing block computes the channel-mixed representation $\mathbf{G}$ by
\begin{equation}
    \mathbf{G}=\operatorname{DiT}\big(\operatorname{concat}(\{\mathbf{f}_d\}_{d=1}^{D},\mathbf{x}_{1:H}^{t})\big)
\end{equation}
where $\mathbf{G}$ 
% denotes the channel-mixed representation produced by the diffusion Transformer block and 
is subsequently fed to the output decoder for step-wise noise regression, $D$ is the number of channels, $\mathbf{f}_d=\operatorname{MLP}(\mathbf{x}_{d,-L+1:0})$ is the per-channel embedding, and $\operatorname{DiT}(\cdot)$ denotes a diffusion Transformer block that aggregates cross-channel temporal modes via multi-head self-attention. 
MCD-TSF \cite{su2025multimodal} incorporates multimodal conditions such as text or spatial descriptors and fuses them with the historical series through cross-attention modules that generate conditioning features for the denoiser. A compact formulation reuses the cross-attention operator above with queries from the time series tokens and keys and values from multimodal tokens.
}

\subsection{Diffusion-centric Condition}
{
Diffusion-centric approaches modify the forward process and the reverse process of diffusion models for time series. We categorize these approaches into two groups, namely \textit{Enhanced process} and \textit{Reformulated process}. The first group injects time series priors into the forward trajectory and reuses these priors in the reverse step. The second group redesigns the forward dynamics to better match the sequence structure.
}

\textbf{Enhanced process} 
{
Real-world time series present time-varying uncertainty due to external factors, whereas the standard diffusion endpoint is a fixed Gaussian $\mathcal{N}(\mathbf{0},\mathbf{I})$ shared by all time series. 
To address this mismatch, this group integrates historical priors into the forward noising trajectory and conditions the reverse step on the same priors.
For example, TMDM \cite{li2024transformer} integrates a Transformer to extract historical features $\hat{\mathbf{y}}_{1:H}$ from $\mathbf{x}_{-L+1:0}$, utilizing this as a conditional prior in both diffusion processes. 
The modified forward process interpolates between the future time series $\mathbf{x}_{1:H}^0$ and the conditional prior $\hat{\mathbf{y}}_{1:H}$ as
\begin{equation}
    \mathbf{x}_{1:H}^t = \sqrt{\alpha_t} \mathbf{x}_{1:H}^0 + (1 - \sqrt{\alpha_t}) \hat{\mathbf{y}}_{1:H} + \sqrt{1 - \alpha_t} \epsilon
\end{equation}
In TMDM, the endpoint distribution at diffusion step $T$ is $\mathcal{N}(\hat{\boldsymbol{y}}_{1: H}, \mathbf{I})$. 
The reverse process employs the prior $\hat{\mathbf{y}}_{1:H}$ in the denoising function $\epsilon_\theta(\cdot)$ and calculate the $\mathbf{X}_{1:H}^{t-1}$ by
\begin{equation}
\begin{split}
    &\mathbf{X}_{1:H}^{t-1} = \\&\frac{1}{\sqrt{\bar\alpha_t}} \left( \mathbf{x}_{1:H}^t - (1-\sqrt{\bar\alpha_t} )\hat{\mathbf{y}}_{1:H} - {\sqrt{1 - \bar\alpha_t}} \epsilon_\theta(\mathbf{x}_{1:H}^t, \hat{\mathbf{y}}_{1:H}, t) \right)
\end{split}
\end{equation}
Then, the prior latent variable $\mathbf{x}_{1:H}^{t-1}$ is obtained by
\begin{equation}
    \mathbf{x}_{1: H}^{t-1}=\gamma_{0} \mathbf{x}_{1:H}^{t}+\gamma_{1} \mathbf{x}_{1: H}^{t}+\gamma_{2} \hat{\mathbf{y}}_{1: H}+\sqrt{\tilde{\boldsymbol{\beta}}_{t}} \boldsymbol{\varepsilon}
\end{equation}
where $\gamma_{0}$, $\gamma_{1}$, $\gamma_{2}$ and $\tilde{\boldsymbol{\beta}}_{t}$ are parameters that depend only on $\beta_t$.
This joint conditioning strategy leverages historical dependencies to refine uncertainty estimation in prediction.
Building on TMDM, NsDiff \cite{ye2025non} further incorporates a historical-dependent variance to model non-stationary uncertainty. The forward process transitions to an endpoint distribution, which is formulated by
\begin{equation}
    \mathcal{N}\big(\mathbf{f}(\mathbf{x}_{-L+1:0}), \mathbf{g}(\mathbf{x}_{-L+1:0})\big)
\end{equation}
where $\mathbf{f}(\cdot)$ and $\mathbf{g}(\cdot)$ are the conditional expected mean function and conditional variance function, respectively. 
The forward process integrates time-varying variances by
\begin{equation}
\begin{split}
    \mathbf{x}_{1:H}^t = \sqrt{\bar{\alpha}_t} \mathbf{x}_{1:H}^0 + (1 - \sqrt{\bar{\alpha}_t}) \mathbf{f}(\mathbf{x}_{-L+1:0})\\ + \sqrt{(\bar{\beta}_t - \tilde{\beta}_t) \mathbf{g}(\mathbf{x}_{-L+1:0}) + \tilde{\beta}_t \sigma_{\mathbf{x}_{1:H}}} \epsilon
\end{split}
\end{equation}
where $\bar{\beta}_t$ and $\tilde{\beta}_t$ are parameters that depend on $\alpha_t$ and $\sigma_{\mathbf{x}_{1:H}}$ is the actual variance. 
The reverse process conditions denoising on both $\mathbf{f}(\mathbf{x}_{-L+1:0})$ and $\mathbf{g}(\mathbf{x}_{-L+1:0})$ to refine uncertainty estimation. 
This approach adapts the diffusion trajectory to temporal uncertainty shifts, improving distributional flexibility.
Besides modeling uncertainty, another study integrates recurrent fusion of historical priors within the forward diffusion process to enhance the utilization of recent temporal information. 
REDI \cite{zhou2024redi} implements a recurrent diffusion mechanism where each forward step combines the current noisy future time series value $\mathbf{x}_i^t$ at time step $i$ with the immediate historical value $\mathbf{x}_{i-1}^t$ before noise addition: 
\begin{equation}
    \mathbf{x}_i^{t} = \sqrt{\alpha_t} (\mathbf{x}_i^{t-1} \oplus \mathbf{x}_{i-1}^{t-1}) + \sqrt{1 - \alpha_t} \epsilon
\end{equation}
where $\oplus$ denotes element-wise addition.
The corresponding reverse process denoises $\mathbf{x}_i^{t}$ to $\mathbf{x}_i^{t-1}$ using the denoised prior $\mathbf{x}_{i-1}^{t-1}$ for step-aware conditioning: 
\begin{equation}
    \mathbf{x}_i^{t-1} = \mu_\theta(\mathbf{x}_i^{t}, \mathbf{x}_{i-1}^{t-1}, t) + \frac{1-\bar{\alpha}_{t-1}}{1-\bar{\alpha}_{t}} \beta_{t} \epsilon
\end{equation}
This recurrent formulation strengthens focus on proximate history to mitigate distribution drift and ensures strict temporal causality by preventing information leakage from future states during generation.
MG-TSD \cite{fan2024mg} and mr-Diff \cite{shen2024multi} adopt multi-scale forward guidance. They build scale-specific priors $\{\hat{\mathbf{y}}_{1:H}^{(s)}\}_{s=1}^{S}$ by downsampling or smoothing the history and aligning the result to the horizon length. The forward step at scale $s$ interpolates toward the corresponding prior through the following process:
\begin{equation}
\mathbf{x}_{1:H}^{t,(s)}=\sqrt{\alpha_t}\,\mathbf{x}_{1:H}^{0,(s)}+(1-\sqrt{\alpha_t})\,\hat{\mathbf{y}}_{1:H}^{(s)}+\sqrt{1-\alpha_t}\,\epsilon
\end{equation}
where $\mathbf{x}_{1:H}^{t,(s)}$ denotes the horizon state at diffusion step $t$ under scale $s$. The hierarchy proceeds from coarse to fine. The reverse step at scale $s$ conditions on $\hat{\mathbf{y}}_{1:H}^{(s)}$ and on the upsampled coarse estimate $U^{(s)}(\tilde{\mathbf{x}}^{(s-1)})$ to refine the current resolution, which is formulated by
\begin{equation}
\mathbf{x}_{1:H}^{t-1,(s)}=\mu_{\theta}\!\big(\mathbf{x}_{1:H}^{t,(s)},\,\hat{\mathbf{y}}_{1:H}^{(s)},\,U^{(s)}(\tilde{\mathbf{x}}^{(s-1)}),\,t\big)
\end{equation}
where $U^{(s)}(\cdot)$ denotes an upsampling operator and $\tilde{\mathbf{x}}^{(s-1)}$ denotes the latest coarse estimate. This coarse-to-fine schedule injects multi-scale priors into the forward trajectory and reuses them in the reverse refinement.
}

\textbf{Reformulated process} 
{
This group replaces the standard noise addition with alternative forward dynamics that reflect temporal structure. Building on the above, this line of studies develops new forward formulations that capture the inherent characteristics of time series.
For instance, DYffusion \cite{ruhling2023dyffusion} leverages the temporal dynamics in the data, directly coupling it with the diffusion steps.
Specifically, this approach designs a stochastic temporal interpolator $I(\cdot)$ for the forward process to generate intermediate states between the initial condition $\mathbf{x}_0$ and the target future point $\mathbf{x}_H$, formalized as minimizing $\mathbb{E} \left\| I_\phi(\mathbf{x}_0, \mathbf{x}_H, i) - \mathbf{x}_i \right\|^2$ where $i$ is an intermediate timestep.
The reverse process employs a forecaster $F(\cdot)$ to predict $\mathbf{x}_H$ from interpolated states, defined as minimizing $\mathbb{E} \left\| F(\mathbf{x}_i, i) - \mathbf{x}_H \right\|^2$, which enhances computational efficiency and enables flexible continuous-time sampling trajectories.
S2DBM \cite{yang2024series} proposes a Brownian Bridge diffusion process to replace the traditional Gaussian diffusion framework. 
This approach constructs a stochastic bridge between a historical prior $\mathbf{h} = F_p(\mathbf{x}_{-L+1:0})$ and the target future $\mathbf{x}_{1:H}^0$, formalizing the forward process as
\begin{equation}
    \mathbf{x}_{1:H}^t = \hat{\alpha}_t \mathbf{x}_{1:H}^0 + (1 - \hat{\alpha}_t) \mathbf{h} + \sqrt{2 \hat{\alpha}_t (1 - \hat{\alpha}_t)} \boldsymbol{\epsilon}
\end{equation}
where $F_p(\cdot)$ denotes the prior predictor and $\hat{\alpha}_t = 1 - t/T$. The reverse process initializes at $\mathbf{x}_{1:H}^T = \mathbf{h}$ and iteratively refines predictions using the following process:
\begin{equation}
    \mathbf{x}_{1:H}^{t-1} = \kappa_t \mathbf{x}_{1:H}^t + \lambda_t \mathbf{x}_\theta(\mathbf{x}_{1:H}^t, \mathbf{h}, \mathbf{c}, t) + \zeta_t \mathbf{h} + \hat{\sigma}_t \epsilon
\end{equation}
where $\kappa_t, \lambda_t, \zeta_t$ are analytical coefficients and $\hat{\sigma}_t^2$ controls variance. 
This architecture integrates historical information as a fixed endpoint to strengthen temporal dependencies and offers configurable noise levels for deterministic or probabilistic forecasting.
ARMD \cite{gao2025auto} reinterprets the diffusion paradigm, treating the future time series $\mathbf{x}_{1:T}^0$ as the initial state and the historical time series $\mathbf{x}_{-T+1:0}$ as the final state\footnote{{In ARMD, the length of the history window and the horizon are the same and equal to the diffusion step length, that is, $L=H=T$.}}. 
Its forward process employs a sliding interpolation technique to generate intermediate sequences $\mathbf{x}_{1-t:T-t}^t$, defined as 
\begin{equation}
    \mathbf{x}_{1-t:T-t}^t =  \sqrt{\bar{\alpha}_t} \mathbf{x}_{1:T}^0 + \sqrt{1 - \bar\alpha_t} \mathcal{S}_w(\mathbf{x}_{1:T}^0,t)
\end{equation}
where $\alpha_t$ decreases with $t$, and $\mathcal{S}_w(\cdot)$ is a evolution trend extraction function that is formulated as
\begin{equation}
    \mathcal{S}_w(\mathbf{x}_{1:T}^0,t)=\left(\sqrt{\frac{1}{\bar{\alpha}_{t}}} \mathbf{x}_{1-t: T-t}^{t}-\mathbf{x}_{1: T}^{0}\right) / \sqrt{\frac{1}{\bar{\alpha}_{t}}-1}
\end{equation}
The reverse process starts from $\mathbf{x}_{-T+1:0}^H = \mathcal{S}_w(\mathbf{x}_{-T+1:0})$ and predicts $\mathbf{x}_{2-t:T-t+1}^{t-1}$ through the following process:
\begin{equation}
    \mathbf{x}_{2-t: T-t+1}^{t-1}=\sqrt{\bar{\alpha}_{t-1}} \hat{X}^{0}_\theta\left(\mathbf{x}^{t}_{1-t:T-t}, t\right)+\sqrt{1-\bar{\alpha}_{t-1}-\sigma_{t}^{2}} \hat{z}_t
\end{equation}
where $\hat{X}^{0}_\theta(\cdot)$ is the devolution network and $\hat{z}_t$ is the predicted evolution trend. This formulation inherently embeds historical information as the generative endpoint, ensuring the reverse trajectory learns a direct mapping from the history to the future, enhancing temporal dependence learning and facilitating sequence modeling.
}

\section{Datasets}
\label{sec:dataset}
{
The datasets used for evaluating diffusion-based TSF models are broadly categorized into two types: unimodal time series datasets and multimodal time series datasets.
% 单变量持续跟踪单个变量，多变量追踪多个变量
Unimodal time series datasets consist solely of temporal signals, where all available information is encoded in time-indexed numerical sequences. Multimodal datasets integrate time series with other modalities such as text, images, or metadata, enabling richer contextual modeling. Table \ref{tab:data_examples_unimodal} presents representative examples of unimodal time series data, while Table \ref{tab:data_examples_multimodal} presents representative examples of multimodal time series data. Table \ref{tab:benchmark} summarizes the benchmark datasets and evaluation metrics used in each diffusion-based TSF approach.
}
\begin{table*}[t]
\centering
\caption{{Examples of unimodal time series datasets covering univariate, multivariate, and spatio-temporal types. For the Traffic dataset, a univariate sequence is extracted from the multivariate records by selecting Sensor 0. For the ETTh1 dataset, three representative variables are plotted to illustrate its multivariate structure. For the SST dataset, the temporal series is drawn at four spatial grid points, each denoted by latitude and longitude coordinates.}}
\label{tab:data_examples_unimodal}
\renewcommand\arraystretch{1.3}
\scriptsize
\begin{tabularx}{\textwidth}{
  >{\centering\arraybackslash}m{1cm}   % 
  >{\centering\arraybackslash}m{1.5cm} %
  >{\centering\arraybackslash}X        %
}
\toprule
\textbf{Type} & \textbf{Datasets} & \textbf{Example} \\
\midrule

\textbf{Univariate}
& Traffic
& \begin{minipage}[c][4cm]{\linewidth}
  \centering
  \includegraphics[width=\linewidth]{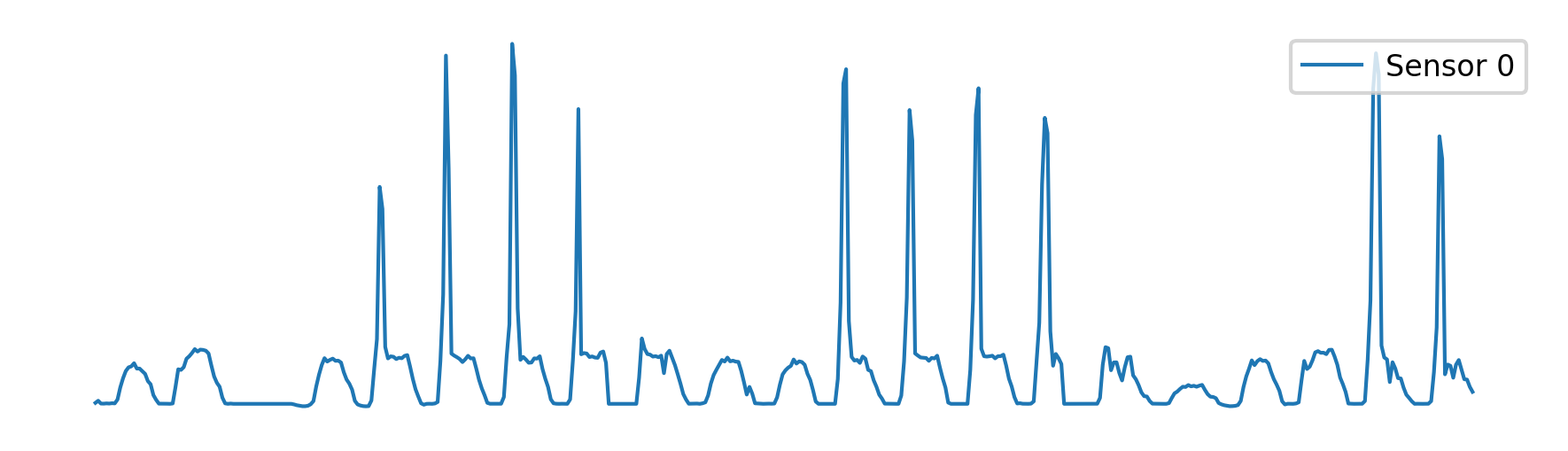}
\end{minipage} \\

\cmidrule(lr){1-3}

\textbf{Multivariate}
& ETTh1
& \begin{minipage}[c][4cm]{\linewidth}
  \centering
  \includegraphics[width=\linewidth]{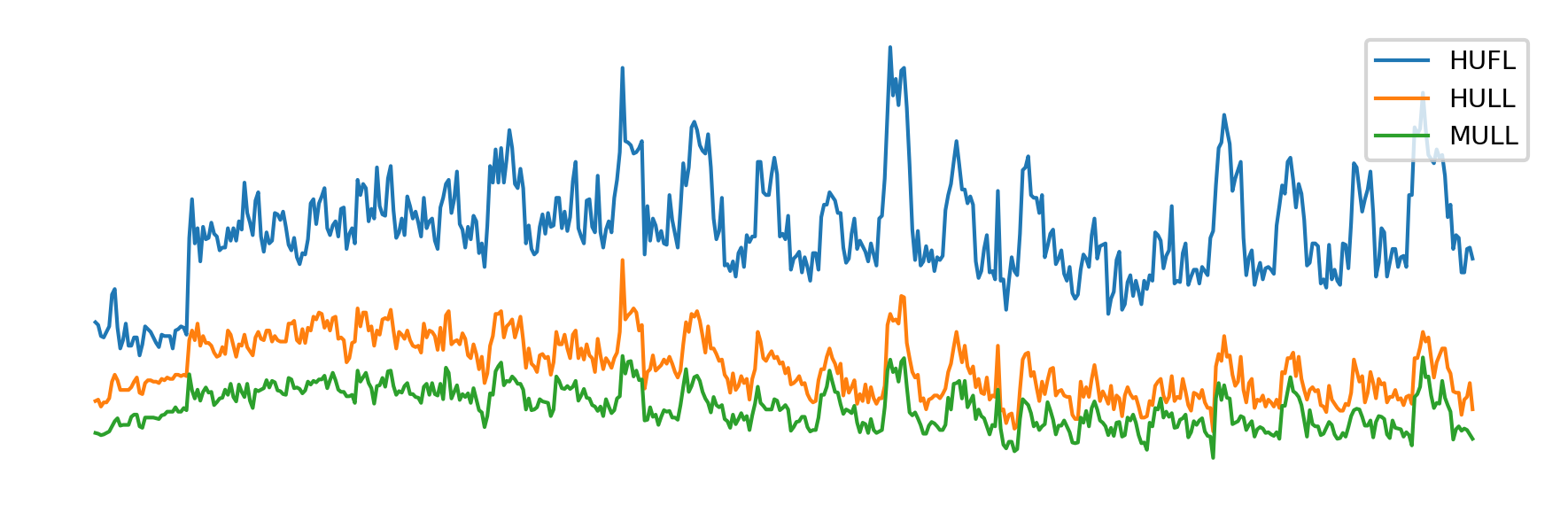}
\end{minipage} \\

\cmidrule(lr){1-3}

\textbf{Spatio-temporal}
& SST
& \begin{minipage}[c][4cm]{\linewidth}
  \centering
  \includegraphics[width=\linewidth]{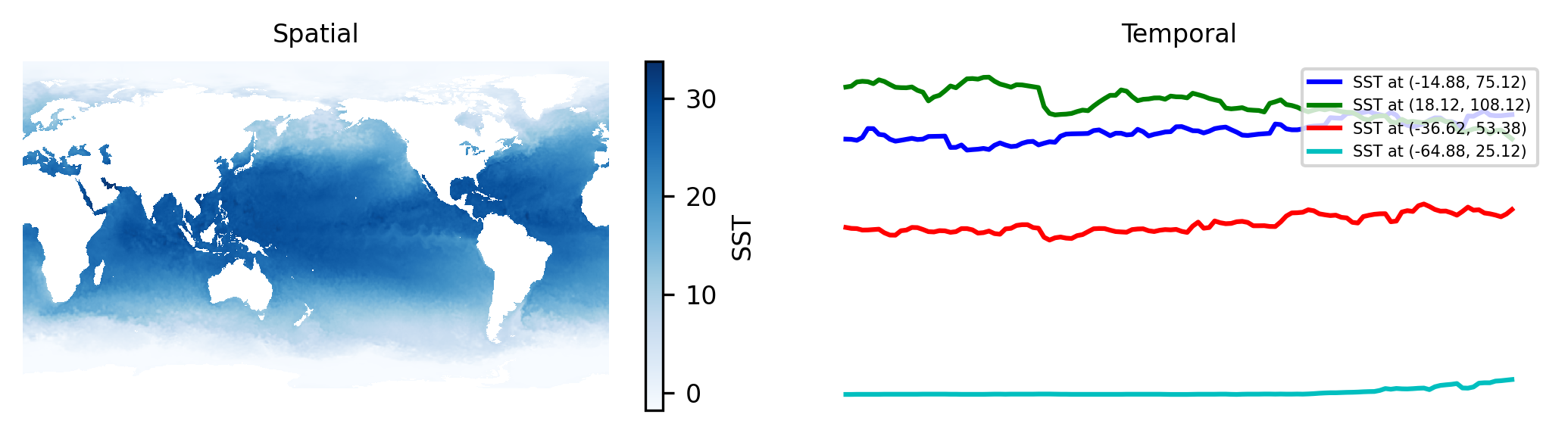}
\end{minipage} \\

\bottomrule
\end{tabularx}
\end{table*}

\begin{table*}[t]
\centering
\caption{{Examples of multimodal TSF datasets. The TTC dataset provides temporally aligned pairs where each timestamp contains both multivariate time series values and their corresponding textual description. The Image-EEG dataset aligns neural time series signals with visual stimuli, where each time series variable is associated with a corresponding image.}}
\label{tab:data_examples_multimodal}
\renewcommand\arraystretch{1.3}
\scriptsize
\begin{tabularx}{\textwidth}{
  >{\centering\arraybackslash}m{1cm}   % 
  >{\centering\arraybackslash}m{1.5cm} % 
  >{\centering\arraybackslash}X        % 
}
\toprule
\textbf{Type} & \textbf{Datasets} & \textbf{Example} \\
\midrule

\textbf{Multimodal}
& TTC
& \begin{minipage}[c][5cm]{\linewidth}
  \centering
  \includegraphics[width=\linewidth]{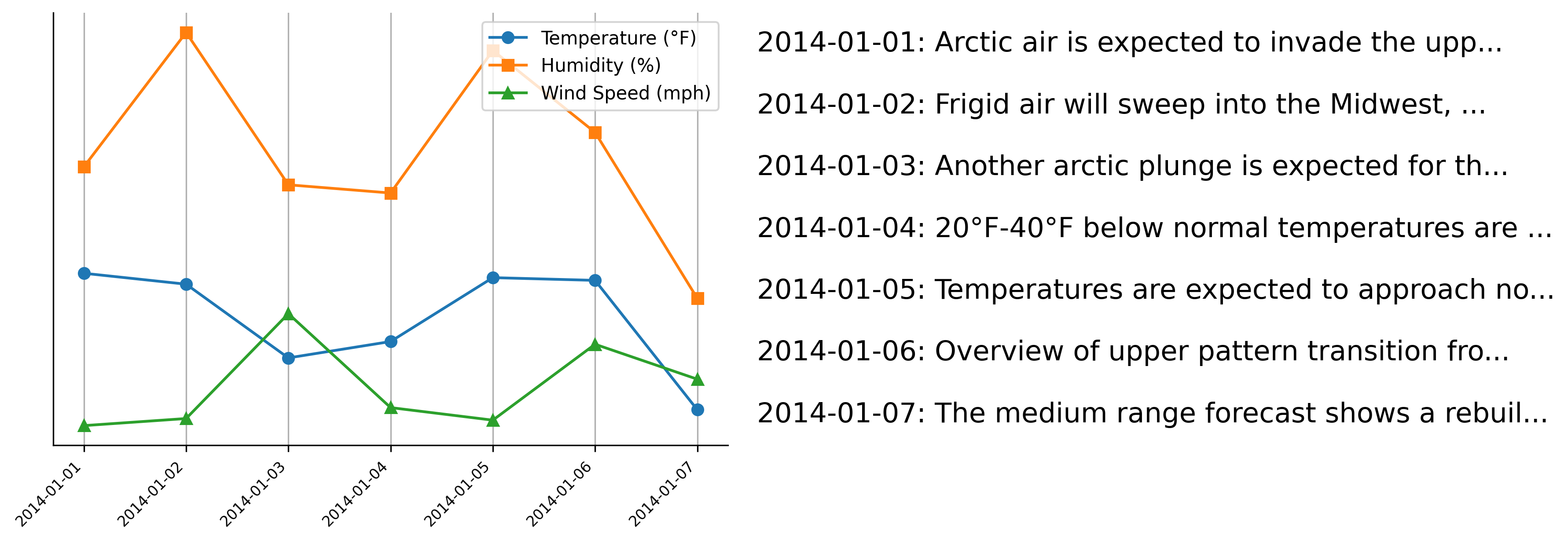}
\end{minipage} \\

\cmidrule(lr){1-3}

\textbf{Multimodal}
& Image-EEG
& \begin{minipage}[c][4.9cm]{\linewidth}
  \centering
  \includegraphics[width=\linewidth]{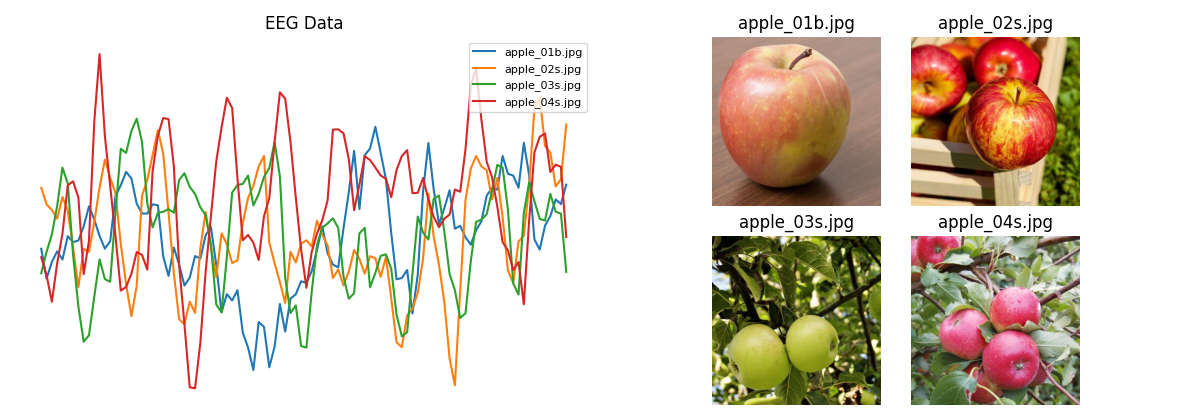}
\end{minipage} \\

\bottomrule
\end{tabularx}
\end{table*}
%--------------------------------------------------- 

\begin{table*}[t]
\caption{{Summary of benchmark TSF datasets and evaluation metrics employed in diffusion models.}}
\label{tab:benchmark}
\small
\centering
\begin{tabularx}{\textwidth}{l|X|r}
\toprule
\textbf{Approaches} & \textbf{Datasets} & \textbf{Evaluation Metrics} \\
\midrule
CSDI \cite{tashiro2021csdi} & Solar, Electricity, Traffic & CRPS \\
\midrule
SSSD \cite{alcaraz2022diffusion} &  Electricity, ETTm1 & MSE, MAE, RMSE, MRE\\
\midrule
DYffusion \cite{ruhling2023dyffusion} & Navier-Stokes Flow & CRPS, MSE \\
\midrule
TimeDiff \cite{shen2023non} & Weather, ETTm1, Traffic, Electricity, ETTh1, Exchange Rates & MSE \\
\midrule
MG-TSD \cite{fan2024mg} & Solar Energy, Electricity, Traffic & CRPS, NMAE, NRMSE \\
\midrule
TMDM \cite{li2024transformer} & Exchange Rates, ILI, ETTm2, Electricity, Traffic, Weather & CRPS, MAE, MSE \\
\midrule
mr-Diff \cite{shen2024multi} & Traffic, Electricity, Weather, Exchange Rates, ETTh1, ETTm1 & MAE \\
\midrule
LDT \cite{feng2024latent} & Solar Energy, Electricity, Traffic & CRPS, MSE \\
\midrule
StochDiff \cite{liu2024stochastic} & Exchange Rates, Weather, Electricity, Solar Energy & NRMSE, CRPS \\
\midrule
Bim-Diff \cite{wang2024treating} & Electricity, Weather, Exchange, ETTh1, ETTm1  & MAE \\
\midrule
CCDM \cite{li2024channel} & ETTh1, Exchange Rates, Weather, Electricity, Traffic & MSE, CRPS \\
\midrule
FDF \cite{zhang2024fdf} & ETTh1, ETTh2, ETTm1, ETTm2, Electricity, Exchange Rates, Weather, Wind & MSE, MAE \\
\midrule
REDI \cite{zhou2024redi} & Weather, Exchange Rates & MAE, RMSE \\
\midrule
RATD \cite{liu2024retrieval} & Exchange Rates, Electricity, Weather & MSE, MAE, CRPS \\
\midrule
S2DBM \cite{yang2024series} & ETTh1, ETTh2, ETTm1, ETTm2, ILI, Weather, Exchange Rates & MSE, MAE \\
\midrule
ARMD \cite{gao2025auto} & Metric, Solar Energy, ETTh1, ETTh2, ETTm1, ETTm2, Exchange Rates & MSE, MAE \\
\midrule
D3U \cite{li2025diffusion} & ETTm1, ETTm2, Weather, Solar Energy, Electricity, Traffic & MSE, MAE \\
\midrule
LDM4TS \cite{ruan2025vision} & ETTh1, ETTh2, ETTm1, ETTm2, Weather, Electricity, Traffic & MSE, MAE \\
\midrule
MCD-TSF \cite{su2025multimodal} & Time-MMD (Agr, Cli, Eco, Ene, Env, Hea, Soc, Tra) & MSE, MAE \\
\midrule
NsDiff \cite{ye2025non} & ETTh1, ETTh2, ETTm1, ETTm2, Electricity, Exchange Rates, ILI, Solar, Traffic &MSE, MAE, CRPS \\
\midrule
FALDA \cite{wang2025effective} & ILI, Exchange, ETTm2, Weather, Electricity, Traffic & MSE, MAE, CRPS \\
\midrule
Diffusion-TS \cite{yuan2024diffusion} & ETTh, Solar Energy & MAE, MSE \\
\bottomrule
\end{tabularx}
\end{table*}

\subsection{Unimodal Time Series Datasets}
{
Unimodal time series datasets refer to those where the input consists of purely temporal numerical data, without involving additional modalities such as text or images. These datasets are further categorized into univariate\footnote{{Univariate time series are constructed by decomposing a multivariate series into separate dimensions, each for an individual variable.}}, multivariate, and spatio-temporal types. Univariate sequences track the evolution of a single variable over time, while multivariate ones involve multiple variables with temporal and cross-variable dependencies. In practice, most datasets are inherently multivariate, and univariate sequences are often constructed by selecting a single variable from them. Spatio-temporal sequences further capture both temporal dynamics and spatial dependencies across locations. We summarize several commonly used unimodal time series datasets below.
}

\subsubsection{Univariate and Multivariate}
\begin{itemize}
\item \textbf{Exchange Rates}\footnote{\url{https://github.com/datasets/exchange-rates/tree/main}}\cite{lai2018modeling}: {A daily multivariate dataset consisting of the exchange rates of eight foreign currencies relative to US dollars. It captures long-term trends and cross-currency correlations, making it suitable for the TSF task in the financial domain.}
\item \textbf{Weather}\footnote{\url{https://www.bgc-jena.mpg.de/wetter/}}: {A multivariate dataset collected from 21 weather stations in Germany, containing hourly meteorological variables such as temperature, humidity, and wind speed. It is widely used for evaluating the model performance for long-range TSF.
}
\item \textbf{Electricity}\footnote{\url{https://archive.ics.uci.edu/ml/datasets/ElectricityLoadDiagrams20112014}}: {A multivariate dataset recording the hourly electricity consumption of 370 clients. Due to its regular seasonality and correlation across variables, it serves as a benchmark for probabilistic and multistep forecasting.
}
\item \textbf{Traffic}\footnote{\url{https://pems.dot.ca.gov./?dnode=Clearinghouse}}: {A multivariate dataset from the {Caltrans Performance Measurement System (PeMS)}\footnote{\url{https://pems.dot.ca.gov/}}, containing 10-minute interval occupancy rates from 963 traffic sensors in California. It exhibits strong daily patterns and is often used for training and evaluating approaches to modeling temporal dependencies across spatial sensors.
}
\item \textbf{Solar Energy}\footnote{\url{https://www.nrel.gov/grid/solar-power-data.html}}: {A multivariate dataset consisting of hourly photovoltaic power production data from 137 solar plants in Alabama. It presents regular day–night cycles and weather-induced variability.
}
\item \textbf{ILI (Influenza-Like Illness)}\footnote{\url{https://gis.cdc.gov/grasp/fluview/fluportaldashboard.html}}: {A multivariate weekly dataset of U.S. outpatient visits due to influenza-like illness, published by the CDC. It is widely used for epidemic forecasting and seasonal modeling.}
\item \textbf{ETT (Electricity Transformer Temperature)}\footnote{\url{https://github.com/zhouhaoyi/Informer2020}}\cite{zhou2021informer}: {A multivariate dataset capturing the hourly oil temperature of electricity transformers and related external features (e.g., load, ambient temperature). It contains multiple variants (ETTh1, ETTh2, ETTm1, ETTm2) differing in frequency and span.
}
\item \textbf{M3 (Makridakis Competition 3)}\footnote{\url{https://github.com/jordicolomer/m3-competition/blob/master/M3Forecast.xls}}\cite{makridakis2000m3}: {A benchmark collection of 3,003 univariate time series from the third Makridakis forecasting competition (M3), covering yearly, quarterly, monthly, and other frequencies across domains such as finance, economics, and industry. It remains widely used for evaluating univariate forecasting accuracy and generalization ability.
}
\end{itemize}

\subsubsection{Spatio-temporal Datasets}
\begin{itemize}
\item \textbf{PEMS04 and PEMS08}\footnote{\url{https://pems.dot.ca.gov/}}\cite{guo2019attention}: {Collected by CalTrans via the Performance Measurement System (PEMS) \cite{chen2001freeway}, these traffic datasets record 5-minute interval flow and speed data from loop detectors. PEMS04 contains readings from 307 sensors in California District 04 (Jan–Feb 2018), while PEMS08 includes data from 170 sensors in District 08 (July–Aug 2018).
}
\item \textbf{Sea Surface Temperature (SST)}\footnote{\url{https://www.psl.noaa.gov/data/gridded/data.noaa.oisst.v2.highres.html}}\cite{huang2021improvements}: A daily-resolution global sea surface temperature dataset from NOAA OISSTv2. This high-resolution SST dataset provides daily, weekly, and monthly SST and sea ice data. As part of NOAA's High-Resolution Blended Analysis, it features a 0.25$^\circ$ global grid covering 89.875$^\circ$S–89.875$^\circ$N and 0.125$^\circ$E–359.875$^\circ$E. The record spans September 1981 to August 2025 and includes a long-term monthly climatology (1991–2020) for reference.
\item \textbf{Navier–Stokes Flow}\footnote{\url{lhttps://archive.nyu.edu/handle/2451/63285}}\cite{otness2021extensible}: {A synthetic fluid dynamics dataset simulating incompressible 2D flows on a 221×42 grid. Each instance includes randomly placed circular obstacles, with x/y velocity fields and pressure channels. It serves as a controlled benchmark for physical simulation and partial differential equation (PDE) forecasting.
}
\item \textbf{Spring Mesh}\footnote{\url{lhttps://archive.nyu.edu/handle/2451/63285}}\cite{otness2021extensible}: {A synthetic physics dataset consisting of a 10×10 particle grid connected by springs. Each node has two positional and two momentum channels. The system evolves based on spring dynamics and provides a clean testbed for TSF approaches with structure-aware sequence modeling.
}
\end{itemize}

\subsection{Multimodal Datasets}
\begin{itemize}
    \item \textbf{Time‑MMD (Multi‑Domain Multimodal Dataset)}\footnote{\url{https://github.com/AdityaLab/Time-MMD/}}\cite{liu2024time}: {The first large-scale multimodal time series dataset spanning nine diverse domains (e.g., agriculture, climate, energy, and traffic). Each sample includes an aligned numerical time series and a fine-grained textual context.}
    \item \textbf{TimeCAP (Time‑Series Contextualization and Prediction)}\footnote{\url{https://github.com/geon0325/TimeCAP}}\cite{lee2025timecap}: {A framework where LLMs serve as both contextualizers and predictors. The dataset includes real-world time series across domains (e.g., weather in New York), paired with summaries generated by GPT-4 \cite{achiam2023gpt} and labeled event outcomes.}
    \item \textbf{TTC (Time‑Text Corpus)}\footnote{\url{https://github.com/Rose-STL-Lab/Multimodal_Forecasting}} \cite{kim2024multi}: {A benchmark comprising temporally aligned pairs of numerical time series and descriptive text in healthcare and climate science.}
    \item \textbf{Image-EEG}\footnote{\url{https://osf.io/nb8wr}}\cite{gifford2022large}: {A large-scale multimodal TSF dataset that pairs time-resolved electroencephalography (EEG) signals with visual stimuli for human object recognition. It includes brain responses from 10 participants viewing over 22,000 images across 30 object categories.}
\end{itemize}

\begin{table*}[t]
\centering
\caption{
{
MSE comparison across experimental settings with different history and prediction lengths. Underlined models denote original sources of reported metrics, while non-underlined entries represent concurrent baselines from those papers. Averages over length combinations are shown in braces \{...\}.
}}
\label{tab: MSE_results}
\renewcommand{\tabularxcolumn}[1]{m{#1}}  %
\newcolumntype{Y}{>{\centering\arraybackslash}X}  %
\begin{tabularx}{\textwidth}{l|*{5}{Y}}
\toprule
\textbf{Dataset} & \textbf{Exchange} & \textbf{Weather} & \textbf{Traffic} & \textbf{Electricity} & \textbf{ETTh1}  \\
\midrule
\textbf{History length} &  \multicolumn{5}{c}{\textbf{\{48,96,192,336\}}}\\
\textbf{Prediction length} &  \multicolumn{5}{c}{\textbf{\{96,168,336,720\}}}\\
\midrule
\underline{CCDM}   & 0.4659 & 0.3414 & 0.8346 & 0.1864 & 0.4769 \\
SSSD  & 0.7349 & 0.4742 & 0.9958 & 0.2532 & 1.0039 \\
TimeDiff        & 0.4522 & 0.4316 & 0.9335 & 0.2048 & 0.4786 \\
TMDM    & 0.8567 & 0.3275 & 0.9364 & 0.2047 & 0.5733 \\
CSDI   & 0.7649 & 0.3065 & 1.4577 & 0.4012 & 1.0642 \\
\midrule
\midrule
\textbf{Dataset} & \textbf{Exchange} & \textbf{Weather} & \textbf{Traffic} & \textbf{Electricity} & \textbf{ETTm2}  \\
\midrule
\textbf{History length} &  \multicolumn{5}{c}{\textbf{\{1000\}}}\\
\textbf{Prediction length} &  \multicolumn{5}{c}{\textbf{\{192\}}}\\
\midrule
\underline{TMDM}   & 0.26 & 0.28 & 0.60 & 0.19 & 0.27 \\
CSDI     & 1.67 & 0.86 &0.94 & 0.56 &1.28 \\
SSSD    & 0.90 & 0.67 &0.81 & 0.47 & 0.97 \\
TimeDiff   & 0.48 & 0.36 & 0.68 & 0.27 & 0.41 \\
\midrule
\midrule
\textbf{Dataset} & \textbf{Exchange} & \textbf{ETTh1} & \textbf{ETTh2} & \textbf{ETTm1} & \textbf{ETTm2}  \\
\midrule
\textbf{History length} &  \multicolumn{5}{c}{\textbf{\{96\}}}\\
\textbf{Prediction length} &  \multicolumn{5}{c}{\textbf{\{96\}}}\\
\midrule
\underline{ARMD}   & 0.093 & 0.445 & 0.311 & 0.337 & 0.181 \\
MG-TSD   & 0.396 & 1.096 &0.295 & 0.690 &0.202 \\
\midrule
\midrule
\textbf{Dataset} & \textbf{Exchange} & \textbf{ETTh1} & \textbf{ETTh2} & \textbf{ETTm1} & \textbf{ETTm2}  \\
\midrule
\textbf{History length} &  \multicolumn{5}{c}{\textbf{\{96\}}}\\
\textbf{Prediction length} & \textbf{14} & \textbf{168} & \textbf{168} & \textbf{192} & \textbf{192}\\
\midrule
\underline{FDF}  & 0.0204 & 0.4364 & 0.3393 & 0.3532 & 0.2303 \\
CSDI  & 1.3023 & 1.1090 &2.0700 & 1.0770 & 1.6051 \\

\bottomrule
\end{tabularx}
\end{table*}

\begin{table*}[t]
\centering
\caption{
{
MAE comparison across experimental settings with different history and prediction lengths. Underlined models denote original sources of reported metrics, while non-underlined entries represent concurrent baselines from those papers. Averages over length combinations are shown in braces \{...\}.
}
}
\label{tab: MAE_results}
\renewcommand{\tabularxcolumn}[1]{m{#1}}  %
\newcolumntype{Y}{>{\centering\arraybackslash}X}  %
\begin{tabularx}{\textwidth}{l|*{6}{Y}}
\toprule
\textbf{Dataset} & \textbf{ILI}& \textbf{Exchange} & \textbf{Electricity} & \textbf{Traffic}  & \textbf{ETTm2} & \textbf{Weather}  \\
\midrule
\textbf{History length} &  \multicolumn{6}{c}{\textbf{\{96\}}}\\
\textbf{Prediction length} &  \multicolumn{6}{c}{\textbf{\{192\}}}\\
\midrule
\underline{FALDA}      & 0.821 & 0.296 & 0.248 & 0.251 &0.301 & 0.255 \\
CSDI      & 1.208 & 0.748 &0.795 & 0.678 & 0.576 & 0.523 \\
SSSD     & 1.079 & 0.861 &0.607 & 0.498 & 0.559 & 0.501 \\
TimeDiff     & 1.085 & 0.429 & 0.690 & 0.851& 0.342 & 0.331 \\
TMDM    & 0.846 & 0.365 & 0.329 & 0.411 & 0.493 & 0.286 \\
D3U     & 0.935 & 0.358 & 0.267 & 0.299 & 0.302 & 0.264 \\
\midrule
\midrule
\textbf{Dataset} & \textbf{ETTh1}& \textbf{ETTh2} & \textbf{ETTm1} & \textbf{ETTm2}  & \textbf{ILI} & \textbf{Solar}\\
\midrule
\textbf{History length} &  \multicolumn{6}{c}{\textbf{\{168\}}}\\
\textbf{Prediction length}& \textbf{192}& \textbf{192} & \textbf{192} & \textbf{192}  & \textbf{36} & \textbf{192}\\
\midrule
\underline{NsDiff}   & 0.594 & 0.514 & 0.488 & 0.281 &2.846 & 0.242 \\
CSDI    & 0.949 & 1.226 &1.002 &1.723 & 4.515 & 0.763 \\
TimeDiff    & 0.517 & 0.456 & 0.537 & 0.268 & 3.958 & 0.821 \\
TMDM      & 0.696 & 0.512 & 0.494 & 0.315 & 3.636 & 0.250 \\
\midrule
\midrule
\textbf{Dataset} & \textbf{Traffic}& \textbf{Electricity} & \textbf{Weather} & \textbf{Exchange}  & \textbf{ETTh1} & \textbf{ETTm1}\\
\midrule
\textbf{History length} &  \multicolumn{6}{c}{\textbf{\{96, 192, 336, 720, 1440\}}}\\
\textbf{Prediction length}& \textbf{168}& \textbf{168} & \textbf{672} & \textbf{14}  & \textbf{168} & \textbf{192}\\
\midrule
\underline{mr-diff} & 0.197 & 0.332 & 0.032 & 0.094 &0.196 & 0.149 \\
CSDI    & 0.468 & 0.540 &0.037 &0.200 & 0.221 & 0.170 \\
TimeDiff     & 0.207 & 0.341 & 0.035 & 0.102 & 0.202 & 0.154 \\
SSSD     & 0.226 & 0.403 & 0.041 & 0.118 & 0.250 & 0.169\\
\bottomrule
\end{tabularx}

\end{table*}

\begin{table*}[t]
\centering
\caption{
{
CRPS comparison across experimental settings with varying history and prediction lengths. Underlined models denote the original source of reported metrics, while non-underlined entries represent concurrent baselines from those papers. Averages over length combinations are shown in braces \{...\}.
}
}
\label{tab: CRPS_results}

\renewcommand{\tabularxcolumn}[1]{m{#1}}
\newcolumntype{Y}{>{\centering\arraybackslash}X}

\begin{tabularx}{\textwidth}{l|*{4}{Y}}
\toprule
\textbf{Dataset} & \textbf{Exchange} & \textbf{Wind} & \textbf{Electricity} & \textbf{Weather} \\
\midrule
\textbf{History length} & \multicolumn{4}{c}{\textbf{\{168\}}} \\
\textbf{Prediction length} & \multicolumn{4}{c}{\textbf{\{96,192,336\}}} \\
\midrule
\underline{RATD}   & 0.339 & 0.673 & 0.373 & 0.301 \\
CSDI    & 0.397 & 0.941 & 0.480 & 0.354 \\
TimeDiff    & 0.589 & 0.917 & 0.490 & 0.410 \\
mr-diff     & 0.397 & 0.881 & 0.429 & 0.347 \\
\midrule
\end{tabularx} 

\begin{tabularx}{\textwidth}{l|*{5}{Y}}
\toprule
\textbf{Dataset} & \textbf{Solar} & \textbf{Electricity} & \textbf{Traffic} & \textbf{taxi} & \textbf{wiki} \\
\midrule
\textbf{History length} & \textbf{168} & \textbf{168} & \textbf{168} & \textbf{48} & \textbf{90} \\
\textbf{Prediction length} & \textbf{24} & \textbf{24} & \textbf{24} & \textbf{24} & \textbf{30} \\
\midrule
\underline{CSDI}  & 0.338 & 0.041 & 0.073 & 0.271 & 0.207 \\
\midrule
\end{tabularx} 

\begin{tabularx}{\textwidth}{l|*{5}{Y}}
\toprule
\textbf{Dataset} & \textbf{ETTh1} & \textbf{ETTh2} & \textbf{ETTm1} & \textbf{ETTm2} & \textbf{ILI} \\
\midrule
\textbf{History length} & \multicolumn{5}{c}{\textbf{\{168\}}} \\
\textbf{Prediction length} & \textbf{192} & \textbf{192} & \textbf{192} & \textbf{192} & \textbf{36} \\
\midrule
\underline{NsDiff}  & 0.392 & 0.358 & 0.346 & 0.256 & 0.806 \\
CSDI     & 0.492 & 0.647 & 0.524 & 0.817 & 1.244 \\
TimeDiff & 0.465 & 0.471 & 0.464 & 0.316 & 1.153 \\
TMDM     & 0.452 & 0.383 & 0.375 & 0.289 & 0.967 \\
\midrule
\end{tabularx}

\begin{tabularx}{\textwidth}{l|*{5}{Y}}
\toprule
\textbf{Dataset} & \textbf{Exchange} & \textbf{Weather} & \textbf{Traffic} & \textbf{Electricity} & \textbf{ETTh1} \\
\textbf{history length} & \multicolumn{5}{c}{\textbf{\{48,96,192,336\}}} \\
\textbf{prediction length} & \multicolumn{5}{c}{\textbf{\{96,168,336,720\}}} \\
\midrule
\underline{CCDM}   & 0.3271 & 0.2563 & 0.3762 & 0.2111 & 0.3600 \\
SSSD     & 0.4666 & 0.3118 & 0.4847 & 0.2630 & 0.5531 \\
TimeDiff & 0.4534 & 0.3526 & 0.5744 & 0.3169 & 0.4418 \\
TMDM     & 0.5144 & 0.2791 & 0.5773 & 0.3163 & 0.4526 \\
CSDI     & 0.5051 & 0.2344 & 0.6104 & 0.3069 & 0.5941 \\
\bottomrule
\end{tabularx}
\end{table*}

\section{Evaluation Metrics}
\label{sec:eval}

{
The performance of a TSF system is evaluated from two complementary perspectives, namely, deterministic metrics and probabilistic metrics.
Deterministic metrics compare a single point forecast to the observed value,
while probabilistic metrics judge an entire predictive distribution.
Tables \ref{tab: MSE_results}, \ref{tab: MAE_results}, and \ref{tab: CRPS_results} present the performance of several diffusion-based TSF approaches on multiple benchmark datasets.
The following subsections illustrate widely used evaluation metrics.
}

\subsection{Deterministic Metrics}
{
Deterministic metrics assess forecasting performance by computing point-wise errors between predicted and ground-truth values.
Among them, Mean Squared Error (MSE) and its variants are widely used due to their sensitivity to large deviations and compatibility with gradient-based learning.
Similarly, Mean Absolute Error (MAE) offers robust alternatives that are less affected by outliers and better suited for evaluating across different scales.
These metrics provide intuitive and interpretable scores, making them a standard choice for deterministic time series forecasting.
In the following {content}, we provide detailed definitions and explanations of each metric.
}

\textbf{Mean Squared Error (MSE)}
{
is computed by
\begin{equation}
    \operatorname{MSE}
    \;=\;
    \frac{1}{H}\sum_{t=1}^{H}
        \lVert x_{t}-\hat x_{t}\rVert^{2}
\end{equation}
where $x_t$ denotes the ground-truth value at time $t$, and $\hat{x}_t$ represents the corresponding forecasted value.
While MSE is easy to compute and differentiable, it penalises large deviations quadratically and is therefore sensitive to outliers.
}

\textbf{Root Mean Squared Error (RMSE)}
{
is computed by
\begin{equation}
    \operatorname{RMSE}
    \;=\;
    \sqrt{\operatorname{MSE}}
\end{equation}
and it is expressed in the same physical units as the observations, making it easier to interpret than MSE.
}

\textbf{Normalized Root Mean Squared Error (NRMSE)}
{
is computed through the following process
\begin{equation}
    \operatorname{NRMSE}
    \;=\;
    \frac{\operatorname{RMSE}}{\sqrt{\frac{1}{H} \sum_{t=1}^{H} \lVert x_t \rVert^2}}
\end{equation}
and it normalizes RMSE by the energy of the true values, yielding a unitless score. 
It facilitates fair comparison between models or datasets with different scales.
}

\textbf{Mean Absolute Error (MAE)}
is computed by
\begin{equation}
    \operatorname{MAE}
    \;=\;
    \frac{1}{H}\sum_{t=1}^{H}
        \lVert x_{t}-\hat x_{t}\rVert_{1}
\end{equation}
where $\ell_{1}$–norm yields a piece-wise linear loss that is more
robust than MSE.

\textbf{Normalized Mean Absolute Error (NMAE)} is computed through the following process
\begin{equation}
    \operatorname{NMAE}
    \;=\;
     \frac{\operatorname{MAE}}{\sum_{t=1}^{H} \lVert x_t \rVert_1}
\end{equation}
It normalizes the absolute error by the magnitude of the ground-truth values, making the metric unitless and scale-invariant.
It is especially useful when comparing performance across datasets or tasks with different value ranges.

\subsection{Probabilistic Metrics}

{
While deterministic metrics overlook uncertainty in predictions, probabilistic metrics assess how well the predicted distribution aligns with the actual observed values.}

\textbf{Continuous Ranked Probability Score (CRPS)}
{
is computed through the following process
\begin{equation}
\operatorname{CRPS}(F, x)
=
\int_{-\infty}^{\infty}
\left(F(z) - \mathbbm{1}\{x \le z\}\right)^2 \, dz
\end{equation}
Here, $\mathbbm{1}\{x \le z\}$ is the indicator function, which equals 1 if $x \le z$ and 0 otherwise. It represents the ground-truth step function centered at $x$. CRPS measures the squared distance between the predicted cumulative distribution function $F(z)$ and the Heaviside step function centered at the ground-truth value $x$. It generalizes MAE to probabilistic forecasts and reduces to MAE when the predicted distribution collapses to a point estimate. In practice, $F$ is approximated by an empirical cumulative distribution function $\widehat{F}_N$, constructed from an ensemble of $N$ sampled predictions $\{\hat{x}^{(n)}\}_{n=1}^{N}$.}

\textbf{Normalized Average CRPS (NACRPS)}
{
To score an $H$-step, $d$-dimensional forecast, NACRPS aggregates the point-wise CRPS values and normalizes them by the total magnitude of the target window:
\begin{equation}
  \operatorname{NACRPS}
  =
  \frac{\displaystyle\sum_{i=1}^{d}\sum_{t=1}^{H}
        \operatorname{CRPS}\!\left(\widehat{F}_{i,t},\, x_{i,t}\right)}
       {\displaystyle\sum_{i=1}^{d}\sum_{t=1}^{H}
        \left|x_{i,t}\right|},
\end{equation}
The $\widehat{F}_{i,t}$ denotes the empirical cumulative distribution function (CDF) for variable $i$ at horizon $t$,
constructed from the ensemble of sampled forecasts. The normalization term rescales the score so that variables with larger magnitudes do not dominate the average, thereby enabling fair comparison across heterogeneous datasets.
}

\begin{figure}[t]
  \centering
  \includegraphics[width=\linewidth]{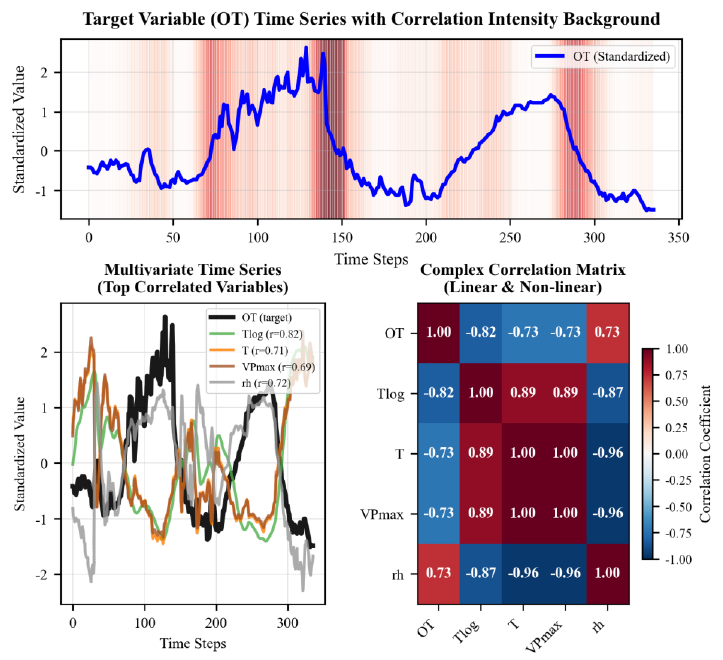}
  \caption{
 The complex dependencies in time series. In the top panel, segments with sharp fluctuations are highlighted with a red background, where darker shades correspond to greater volatility and typically indicate more intricate temporal dynamics. In the bottom-left panel, four variables that are most correlated with the target variable (OT) are overlaid to demonstrate cross-variable coupling. The bottom-right panel presents the pairwise correlation matrix across variables, which reveals complex inter-variable dependencies.
  }
  \label{fig:complex_correlation}
  \vspace{-0.2cm}
\end{figure}

\begin{figure}[t]
  \centering
  \includegraphics[width=\linewidth]{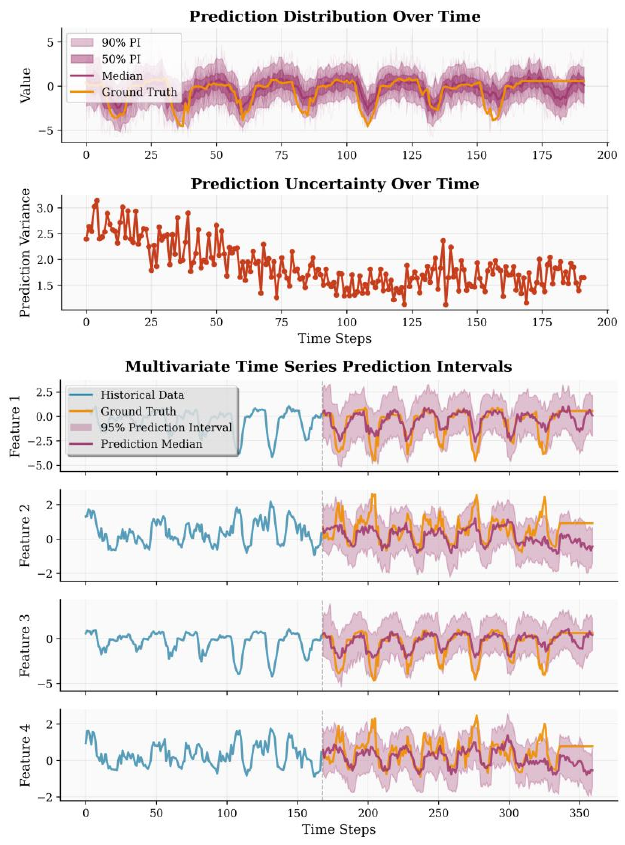}
  \caption{
  {
  Dynamic uncertainty in time series. In the top panel, the evolution of prediction intervals (PIs) and predictive variance highlights the time-varying nature of uncertainty. In the bottom panel, the PIs for four variables in a multivariate time series are shown, demonstrating how uncertainty evolves over time across different variables.
  }
  }
  \label{fig:dynamic_uncertainty}
  \vspace{-0.2cm}
\end{figure}

\section{Discussions}
\label{sec:discussion}
{
In the aforementioned sections, we review the foundation diffusion models and diffusion-based TSF approaches, and then provide a systematic categorization and detailed description of these approaches, focusing on their conditioning sources and integration process. 
In this section, we summarize key characteristics of current work and discuss the limitations and challenges that remain in diffusion-based TSF.
}

\subsection{Overall Analysis of Existing Studies}
{
\textbf{Key characteristics} Diffusion-based TSF establishes a unified generative paradigm that addresses four core needs of temporal prediction. 
First, it models complex dependencies along the time axis and across variables through stepwise denoising with learnable score functions. 
This structure supports long range temporal effects and preserves cross time series interactions during sampling, consistent with the correlation patterns shown in Fig. \ref{fig:complex_correlation}. 
Second, it delivers probabilistic forecasts by drawing trajectories from the learned reverse dynamics, which yields full predictive distributions rather than only point values. 
The resulting intervals vary with regime changes and reflect dependence among variables, in agreement with the variability illustrated in Fig. \ref{fig:dynamic_uncertainty}. 
Third, the conditional formulation provides flexibility for injecting prior knowledge from the history and from auxiliary sources. 
Domain preprocessing separates structure from noise through trend and seasonality decomposition, multiscale analysis, and frequency domain transforms such as Fourier decomposition. 
The extracted components supply explicit conditions that guide central tendency and dispersion during denoising. Extensions broaden the source of conditions with textual context, spatial or relational graphs, and retrieval based references so that external cues complement the history when evidence in the past window is weak. 
Fourth, empirical studies on common benchmarks report competitive accuracy together with calibrated predictive intervals, and robustness often improves when conditions encode strong time series priors and when integration preserves temporal structure throughout the trajectory.
}

{
\textbf{Two-layer perspective across approaches} Existing approaches organize these gains around two coordinated layers. 
The first layer enriches the conditioning source so that the diffusion model receives disentangled temporal features and auxiliary evidence derived from the historical sequence and from complementary modalities. 
Representative choices include decomposition into trend, seasonality, and residual, multiscale summaries that capture coarse to fine patterns, and frequency domain components that describe periodic structure. 
Multimodal additions supply side information that clarifies context when the raw sequence is ambiguous and they expand the prior beyond what the history alone supports. 
The second layer refines the integration pathway so that conditions shape generation in a manner aligned with temporal dynamics and with consistent propagation of uncertainty. 
Some approaches inject conditions through the denoising network to steer the score toward patterns supported by priors, while other approaches embed conditions into the diffusion trajectory by adjusting schedules or latent covariances so that guidance acts within the forward and reverse dynamics. 
This two-layer view links observed advantages in dependency modeling, uncertainty estimation, conditioning flexibility, and benchmark performance to concrete choices on where priors enter and how they interact with the generative process, and it aligns the narrative with Fig. \ref{fig:complex_correlation} and Fig. \ref{fig:dynamic_uncertainty} that motivate the focus on correlations and dynamic uncertainty.
}

\subsection{Limitations and Challenges}
{
\textbf{Model rigidity and generalization} Many diffusion-based TSF approaches bind the forecasting specification to the architecture. 
The history length, the variable set, and the horizon are fixed by the input and output signature and by positional encodings, which restricts adaptation when any of these quantities changes, when channels appear or disappear, or when sampling becomes irregular. 
Structural priors optimized for a single configuration provide limited support for variable rate resampling and for missing segments, and they constrain the conditioning pathway to a narrow family of shapes. 
Retuning for each setting raises operational overhead and weakens reproducibility because hyperparameters and early stopping criteria drift across configurations. 
The absence of shape robust encoders and horizon aware heads also limits cross dataset transfer and out of domain generalization under distribution shift, since the learned score functions specialize to a restricted dependency range and to a fixed cross variable topology. 
The constraint further affects hierarchical forecasting and rolling updates where the specification must adjust on the fly, and it reduces reliability when new sensors appear, when some channels become inactive, or when covariates follow a different sampling regime.
}

{
\textbf{Uncertainty estimation and interpretability} Most approaches produce intervals or full predictive distributions, yet the mapping from interval width to causal factors remains unclear. 
Forecast require explanations of where risk concentrates and why a given step on the horizon receives wider bounds. 
Current pipelines rarely partition uncertainty into aleatoric and epistemic components or attribute variance to trend change, seasonal phase shift, exogenous shocks, or sensor noise. 
Calibration practice often centers on aggregate scores such as CRPS and underreports horizon wise reliability, tail behavior, and dependence across variables. 
Extreme events and regime transitions stress these weaknesses because heavy tailed errors and non stationary volatility receive little attention in the noise schedule or in the structure of the score network. 
Numerical validity therefore provides limited decision value when the intervals lack traceability to concrete signals in the history or in auxiliary inputs. 
In addition, interval sets sometimes violate monotonicity across confidence levels, quantiles cross on long horizons, and uncertainty fails to remain temporally coherent, which reduces trust in downstream risk controls.
}

{
\textbf{Multimodal utilization and temporal asynchrony} Recent approaches extend the conditioning source with text, spatial graphs, images, or external knowledge, yet integration frequently reduces to side feature concatenation to the denoiser. 
Multimodal information rarely shapes the forward diffusion prior or the reverse dynamics, which limits the influence of external context on generation. 
Streaming environments introduce asynchronous arrivals where the time series modality appears first and textual or knowledge signals arrive with latency. 
Without explicit alignment operators the model ignores delayed inputs or attaches them to the wrong forecast step, which injects noise that propagates across the trajectory and degrades both accuracy and calibration. 
Few studies explore asynchronous supervision during training, lag aware conditioning at inference, or memory mechanisms that revise beliefs once a delayed modality becomes available. 
The gap is larger when modalities are intermittently missing or when time stamps drift across data sources, since the alignment error accumulates across the sampling path.
}

{
\textbf{Modality confidence and reliability weighting}
Heterogeneous sources present variable reliability over time. 
Textual reports may contain uncertainty markers, sensor nodes may drift, and retrieved references may mismatch context. 
Existing approaches assign static fusion weights or rely on attention without a calibrated notion of source confidence. 
The absence of confidence aware fusion affects the mean forecast and the spread of the distribution. 
Effective fusion requires reliability signals tied to the diffusion state or to the scheduler so that unreliable modalities receive reduced influence in high variance regimes and authoritative cues receive higher influence during regime transitions, which remains underexplored. 
Failure to manage reliability also hampers detection of modality dropouts and adversarial noise, and it prevents selective abstention when external evidence conflicts with the historical pattern.
}

{
\textbf{Computational efficiency and deployability} Iterative sampling introduces latency that grows with horizon and dimensionality. 
Long sequences demand more denoising steps for stability, and multivariate targets increase memory traffic and compute per step. 
High cost restricts real time deployment and narrows the scope of hyperparameter exploration during development. 
Accelerated solvers and distillation reduce step counts, yet aggressive compression risks calibration drift and distortion of cross variable dependence, which creates a trade off between speed and probabilistic fidelity. 
Systems factors add friction, including suboptimal batching for variable length windows and bandwidth bottlenecks during multistep sampling. 
Training cost also remains high because score models require long schedules, large batches for stable variance estimation, and frequent checkpointing, which limits iteration speed in applied settings.
}

{
\textbf{Evaluation breadth and practical fidelity} Benchmark coverage often concentrates on a limited set of datasets and fixed horizons, which hides sensitivity to configuration changes and to non stationary regimes. 
Many studies prioritize point metrics and a small set of probabilistic scores while omitting diagnostics that reveal horizon wise calibration, tail risk, conditional coverage across contexts, and stability of prediction sets under perturbations. 
Protocols rarely stress asynchronous multimodal input, delayed feedback, or realistic missingness patterns. 
Limited evaluation breadth obscures failure modes and slows progress on the limitations described above. 
Reporting sometimes omits computational budgets, solver step counts, and memory profiles, which complicates fair comparison and makes deployment trade offs less transparent.
}

{
\textbf{Summary of open problems}
The field faces a linked set of obstacles. 
Structural adaptability remains limited by fixed input output shapes and narrow configuration choices, which reduces reliability under specification drift. 
Uncertainty estimation lacks interpretability that traces risk to trend, seasonality, exogenous events, or sensor reliability, and temporal coherence of intervals remains fragile. 
Multimodal fusion stays shallow and overlooks asynchrony and confidence, which reduces the value of external context in streaming environments. 
Computational cost escalates with sequence length and dimensionality, and naive acceleration threatens calibration and cross variable dependence. 
Evaluation practice trails practical needs and does not reflect deployment constraints such as latency budgets and intermittent modalities. 
Addressing these issues requires architectures that adjust structure without full retraining, uncertainty pathways that attribute risk to identified sources, fusion mechanisms that handle delays and reliability, sampling strategies that reduce steps without eroding probabilistic quality, and benchmarks that stress the scenarios encountered in real systems.
}

\section{Future Directions}
\label{sec:direction}
{
Diffusion-based TSF continues to advance along several complementary axes. 
The directions below focus on scalable pretraining, interpretable uncertainty, trajectory-level multimodal fusion, process optimization for temporal structure, reduction of inference cost, and broader application coverage, with an emphasis on concrete mechanisms and evaluation protocols that match real deployment settings.
}

{
\subsection{Foundation diffusion models for universal TSF} A foundation diffusion model for time series pretrains a generative prior on heterogeneous corpora spanning domains, sampling regimes, and horizon specifications. 
The objective covers forecasting under diverse history lengths, variable sets, and horizons, together with multi view masking, reconstruction across gaps, trajectory infilling at multiple scales, and contrastive alignment across datasets. 
Regularization promotes invariance to time shift, amplitude scaling, and calendar effects. Representation sharing separates domain invariant dynamics from domain specific signals so that the prior transfers across tasks with minimal additional training. 
Parameter efficient adaptation supports specialization through lightweight adapters, prompt-based conditioning, and low rank updates that preserve the pretrained prior while aligning to a target dataset. 
Retrieval augmented conditioning supplies similar histories or prototypical patterns from large repositories. Retrieved sequences act as structured conditions that guide score estimation and reduce brittleness under distribution shift and data scarcity. 
Practical deployment benefits from continual pretraining with drift detection, from domain routers that activate experts under new regimes, and from governance of data provenance. A single backbone with these capabilities sustains robustness under drift and reduces per-task retraining.
}

{
\subsection{Explainable uncertainty for risk prediction} Future work on uncertainty targets statistical reliability together with transparent attribution. 
Noise schedules and covariance structures tie to history dependent states and external drivers so that uncertainty tracks volatility cycles, regime shifts, and detected structural breaks. 
Predicted quantiles form time‐indexed curves that remain ordered. At each step the upper quantiles stay above the lower ones and the curves do not cross. Prediction intervals widen as the confidence level increases across the horizon, which preserves internal consistency.
Attribution maps connect interval width to trend change, seasonal phase, exogenous events, and sensor quality, with step wise summaries that trace how risk propagates across the horizon. 
Evaluation report coverage by horizon, by context, and by event type, rather than only aggregate scores. 
Domain knowledge enters through soft constraints on the score or regularization of the scheduler, which suppresses trajectories that violate physical or operational limits. 
The outcome is a forecast that reports values, uncertainty, and reasons, with uncertainty that is interpretable and traceable across time.
}

{
\subsection{Diffusion centric multimodal fusion} Multimodal fusion advances beyond side features by embedding auxiliary signals directly into the diffusion trajectory. 
Text, spatial or relational graphs, images, and knowledge sources influence the forward prior, the reverse drift, and the latent covariance so that external context shapes generation at each step. 
Temporal asynchrony receives explicit treatment through alignment operators that account for lags and variable sampling rates during both training and inference. 
Streaming settings benefit from memory modules that revise beliefs when a delayed modality arrives and from schedules that simulate realistic delays. 
Reliability varies across sources, therefore confidence weighting becomes a first class component that modulates both the mean and the dispersion. 
Confidence signals derive from self-assessment of each modality encoder, from agreement between sources, and from historical calibration of attribution errors. 
Training curricula include lag perturbation, modality dropout, and conflict resolution so that the fusion policy remains stable under missing or noisy inputs. 
This trajectory level design delivers deeper fusion and raises robustness when the history window provides weak evidence.
}

{
\subsection{Process optimization for temporal structure} Forward and reverse dynamics align more closely with properties of time series. 
Schedulers adapt across scales so that coarse trend and fine seasonal components receive appropriate step sizes. 
Decomposition-aware designs route trend, seasonality, and residual through coupled paths that exchange information during denoising. 
Frequency-aware latents represent periodic structure in the spectrum while time domain latents retain local transitions, with cross domain coupling that preserves phase information. 
Time warping maps irregular sampling to a stationary frame and returns forecasts to the original clock, which stabilizes training when intervals vary. 
Causality respecting trajectories restrict information flow to past and present conditions and enforce lag structure across variables. 
These refinements produce trajectories that reflect trend, seasonality, non stationarity, and dependence.
}

{
\subsection{Lowering inference cost} Efficient forecasting with diffusion requires fewer steps and better systems integration. 
Learned solvers and consistency objectives distill long schedules into short sampling paths while preserving calibration and cross variable dependence. 
Horizon-aware step allocation places more steps where dynamics change rapidly and fewer steps where signals remain stable. 
State caching and shared computation across output channels reduce redundant score evaluations. 
Operator fusion, mixed precision, quantization aware training, and memory efficient attention lower latency and memory footprint. 
Early exit rules stop sampling when mean and dispersion stabilize under predefined tolerances with guarantees that respect coverage targets. 
These choices expand the feasible range of horizons and dimensionalities under real time budgets and improve the reproducibility of reported costs.
}

{
\subsection{Application expansion} Diffusion for time series extends to domains with irregular sampling, exogenous shocks, and rich context. 
Finance requires event aware volatility tracking, strict auditability of uncertainty, and stress tests that reflect liquidity and market microstructure. 
Healthcare requires alignment between clinical notes and physiological signals, robust handling of missingness, and safeguards that protect patient privacy. 
Industrial monitoring benefits from graph informed priors over sensor networks, early warnings for regime changes, and attribution that localizes faults. 
Earth observation and renewable energy introduce large scale spatiotemporal structure with external drivers from meteorology and policy. 
Human mobility and operations management require integration of control signals and resource constraints. 
Progress in these areas depends on datasets and protocols that reflect streaming input, delayed modalities, and deployment constraints, together with metrics that assess accuracy, calibration, attribution quality, and decision utility.
}

\bibliographystyle{IEEEtran}
\bibliography{ref}

\newpage

\appendices

\end{document}